\documentclass{article}

\usepackage[letterpaper,margin=3cm,footskip=30pt]{geometry}
\usepackage{amsmath}
\usepackage{amsfonts}
\usepackage{amsthm}
\usepackage{paralist}
\usepackage{booktabs}
\usepackage{graphicx}
\usepackage{array}
\usepackage{multirow}
\usepackage{multicol}
\usepackage{xcolor}
\usepackage{xspace}
\usepackage{hyperref}
\usepackage{balance}
\usepackage[most]{tcolorbox}
\usepackage{caption}
\usepackage{subcaption}
\usepackage{listings}
\usepackage{url}
\usepackage{authblk}
\usepackage{soul}
\usepackage{comment}
\usepackage{bbding}
\usepackage{amsmath}

\definecolor{lightred}{RGB}{255, 200, 200}
\definecolor{lightgreen}{RGB}{200, 255, 200}
\definecolor{deepviolet}{RGB}{96, 0, 96}

\tcbset{
  promptbox/.style={
    top=10pt,
    colback=white,
    colframe=violet,
    colbacktitle=deepviolet,
    fontupper=\itshape,
    enhanced,
    center,
    attach boxed title to top left={yshift=-0.1in,xshift=0.15in},
    boxed title style={boxrule=0pt,colframe=white},
  }
}
\newtcolorbox{prompt}[2][]{promptbox,title=#2,#1}

\newcommand{\myparagraph}[1]{\vspace{1ex}\noindent\underline{\it #1.}\xspace}

\newcommand{\set}[1]{\{#1\}\xspace}

\newcommand{\var}[1]{\texttt{\set{\upshape #1}}}

\title{``Guinea Pig Trials'' Utilizing GPT: A Novel Smart Agent-Based Modeling Approach for Studying Firm Competition and Collusion}

\newcommand{\fordham}{$^{1}$}
\newcommand{\osaka}{$^{2}$}
\newcommand{\kyoto}{$^{3}$}
\newcommand{\nagoya}{$^{4}$}
\newcommand{\osakyo}{$^{2,3}$}
\newcommand{\osanag}{$^{2,4}$}

\author{
{\fordham Xu Han, \osakyo Zengqing Wu, \osanag Chuan Xiao}\\
\small{\fordham Fordham University, \osaka Osaka University, \kyoto Kyoto University, \nagoya Nagoya University}\\
\vspace{1ex}
\small{xhan44@fordham.edu, wuzengqing@outlook.com, chuanx@nagoya-u.jp}
}

\date{}

\begin{document}

\maketitle

\begin{abstract}
    Firm competition and collusion involve complex dynamics, particularly when considering communication among firms. Such issues can be modeled as problems of complex systems, traditionally approached through experiments involving human subjects or agent-based modeling methods. In this study, we propose an innovative framework called Smart Agent-Based Modeling (SABM), wherein smart agents, supported by powerful GPT-4 technologies, represent firms, and interact with one another. We conducted a controlled experiment to study firm price competition and collusion behaviors under various conditions. SABM is more cost-effective and flexible compared to conducting experiments with human subjects. Smart agents possess an extensive knowledge base for decision-making and exhibit human-like strategic abilities, surpassing traditional ABM agents. Furthermore, smart agents can simulate human conversation and be personalized, making them ideal for studying complex situations involving communication. Our results demonstrate that, in the absence of communication, smart agents consistently reach tacit collusion, leading to prices converging at levels higher than the Bertrand equilibrium price but lower than monopoly or cartel prices. When communication is allowed, smart agents achieve a higher-level collusion with prices close to cartel prices. Collusion forms more quickly with communication, while price convergence is smoother without it. These results indicate that communication enhances trust between firms, encouraging frequent small price deviations to explore opportunities for a higher-level win-win situation and reducing the likelihood of triggering a price war. We also assigned different personas to firms to analyze behavioral differences and tested variant models under diverse market structures. The findings showcase the effectiveness and robustness of our SABM framework and provide intriguing insights into competition and collusion. Our source code is available at \url{https://github.com/Roihn/SABM} .
\end{abstract}

\section{Introduction}
\label{sec:intro}
Complex systems encompass natural or social systems consisting of multiple agents, where the behavior of the entire system emerges from interactions among these agents. In the realm of business, numerous topics fall under the category of complex system problems, including customer relationship management, market dynamics, price competition, mergers, and collusions, etc. The emergent properties exhibited by complex systems pose challenges for traditional top-down modeling methods, as they struggle to capture individual-level behaviors effectively. Instead, researchers typically employ ground-up approaches that allow for the modeling of individual behaviors. In this context, behavioral studies, and agent-based modeling (ABM) serve as the primary research methodologies commonly found in the literature on complex systems.

In behavioral studies, human participants are typically enlisted to perform within controlled lab or field experiment settings~\cite{tor2010fairness,fonseca2012explicit,andres2023communication}. Conducting experiments in lab settings, however, is associated with several limitations. Firstly, participants are often aware that they are partaking in an experiment, and the artificial environment of the lab may lack realism. Additionally, the characteristics of the participants recruited for the experiment may deviate from those of the broader population and introduce bias to the results. Another common issue with lab experiments is their limited ecological validity, as it can be challenging to adjust and generalize findings to real-world situations. Conducting field experiments can partially mitigate these limitations and is generally preferred. However, field experiments are often expensive and, in some cases, impractical to carry out.

Agent-based modeling (ABM) is another commonly utilized method for studying complex systems and has demonstrated some success~\cite{tesfatsion2002agent,grimm2005pattern,perez2009agent}. Traditional ABM approaches primarily rely on rule-based models, where explicit rules govern the behavior and decision-making of individual agents. These rules are typically designed by domain experts or derived from empirical data and knowledge, and they remain static throughout the simulation. While rule-based models can capture specific patterns and interactions within the system, they possess certain limitations. They may oversimplify agent behaviors and prove challenging to validate if the rules are subjective or lack empirical support~\cite{janssen2006empirically}. Moreover, rule-based models do not account for learning or adaptation over time. 

In recent years, there has been a surge in the popularity of learning-based ABM methods due to advancements in AI. These models enable agents to learn, adapt, and modify their behavior based on experiences and interactions within the environment. Various approaches, such as reinforcement learning and evolutionary algorithms, can be employed to implement learning-based ABM. They prove particularly valuable when the decision-making processes of agents are not fully understood or when the environment is dynamic and uncertain. However, learning-based ABM models also have their limitations. They require a sufficient amount of data for learning, which may not always be readily available. Designing and implementing learning algorithms can be complex and demand expertise on machine learning techniques. Furthermore, learning-based models can be computationally intensive, especially when training a large number of agents or employing complex learning algorithms. The hardware costs associated with training AI rapidly escalate to a level that becomes prohibitively expensive for the majority of individual researchers and even for universities. 

The recent surge in Generative AI techniques presents new opportunities for addressing complex systems. The concept of Generative Pre-trained Transformer (GPT) was introduced by OpenAI in 2018, emphasizing its ability to generate human-like content and engage in human-like conversations. In November 2022, OpenAI introduced ChatGPT, a Large Language Model (LLM) that was pre-trained and fine-tuned based on the powerful GPT-3.5. ChatGPT not only excels in conversational interactions but also produces high-quality long-form writings. The release of ChatGPT garnered significant public attention and is widely regarded as one of the most groundbreaking technologies of recent decades. While people are still astonished by ChatGPT's remarkable performance, OpenAI introduced GPT-4 on March 14, 2023, which currently stands as the most powerful generative AI available, surpassing similar products. In complex tasks such as the GRE tests or bar exams, GPT-4 surpasses the performance of most human test takers, placing within the top 10\%~\cite{bubeck2023sparks}.

In this paper, we propose an innovative Smart Agent-Based Modeling (SABM) framework that incorporates the GPT-4 technologies for studying complex systems in the business world. In SABM, multiple smart agents, based on GPT-4, are created to interact with each other within a game-theoretical setting. By using smart agents as ``guinea pigs,'' we significantly reduce costs and improve the reproducibility and controllability of experiments, compared to conducting experiments with human subjects. Furthermore, concerns related to safety and ethical issues commonly associated with human subject experiments are eliminated. Compared to traditional ABM, smart agents possess a superior knowledge base and reasoning ability that rivals that of human beings~\cite{brown2020language,binz2023using}. They have the capability to comprehend and generate contextual materials, enabling the simulation of human-like conversations within the SABM framework, a feat not previously accomplished in traditional ABM systems. Additionally, smart agents can be assigned personas, providing intriguing opportunities to investigate the intricacies of human decision-making under various circumstances~\cite{radford2019language}.

We utilize the SABM framework to investigate the dynamics of firms' price competition and collusion formation under different scenarios, considering the presence or absence of communication among firms. The remainder of this paper is structured as follows. In Section~\ref{sec:design}, we outline the specific settings of our price competition game. We then focus on our base model, which prohibits communication, and present the corresponding results in Section~\ref{sec:base}. Moving on to Section~\ref{sec:communication}, we explore an alternative model that allows for communication and compare its outcomes to those of the base model in terms of equilibrium prices, the smoothness of price convergence, and the speed of collusion formation. In Section~\ref{sec:variants}, we examine several intriguing variants of the base model to gain further insights. We conclude our paper in Section~\ref{sec:concl}.

\section{Experiment Design}
\label{sec:design}
We apply the SABM framework in a canonical Bertrand duopoly setting, considering differentiable goods, and conduct an infinitely repeated pricing game. The experimental settings are elaborated in detail as follows.

\subsection{Bertrand Duopoly Game with Differentiable Goods}
We investigate a scenario featuring two firms, referred to as Firm 1 and Firm 2, both offering distinct types of differentiable goods. These goods are substitutable, and the degree of differentiation is determined by specific parameters. Each firm operates as a Bertrand-type firm, engaging in price competition. Firms are profit maximizers, and the information is complete. Each firm has its own demand function $q_i(p_i, p_j)$ where $i, j \in \set{1, 2}$, and $i \neq j$. We assume no Giffen goods, i.e., $\frac{\partial q_i}{\partial p_i} < 0$, and we assume firms have gross substitute valuation, i.e., $\frac{\partial q_i}{\partial p_j} > 0$. Each firm $i$ has a constant marginal cost $c_i$, and its profit is denoted as $\pi_i = (p_i - c_i) q_i$. We assume that $\frac{\partial^2 \pi_i}{\partial p_i^2} \leq 0$ and $\frac{\partial^2 \pi_i}{\partial p_i p_j} \geq 0$, so that a firm's marginal profit goes down with its own price but goes up with its rival's price. We further assume that $\frac{\partial^2 \pi_i}{\partial p_i^2} \cdot \frac{\partial^2 \pi_j}{\partial p_j^2} > \frac{\partial^2 \pi_i}{\partial p_i p_j} \cdot \frac{\partial^2 \pi_j}{\partial p_j p_i}$, which guarantees that a firm's own effects dominate the cross effects. We adopt a linear demand function which easily satisfies all the conditions listed above. The linear inverse demand functions are given as
\begin{align}
    p_1 &= a - \beta q_1 - dq_2, \\
    p_2 &= a - \beta q_2 - dq_1, 
\end{align}
where $d/\beta$ is the parameter controlling the level of differentiation and $d/\beta \in [0, 1]$. If $d/\beta = 1$, the two products are homogeneous and are perfect substitutes. The level of product differentiation increases when $d/\beta \to 0$. When $d/\beta = 0$, the model reduces to the monopoly case where no substitute is available for each product. Parameter $a$ serves as an upper bound for prices when demands are $0$. The linear demand function for each firm is thus derived as follows. 
\begin{align}
    q_1 &= \frac{1}{b} (\alpha - \beta p_1 + dp_2), \\
    q_2 &= \frac{1}{b} (\alpha - \beta p_2 + dp_1), 
\end{align}
where $b = \beta^2 - d^2$, and $\alpha = a\beta - ad$. 

The goal for each firm is to maximize its profit $\pi_i = (p_i - c_i) q_i$, $i \in \set{1, 2}$. The profit for each firm can then be derived and the Bertrand equilibrium prices can be calculated, as shown in the following equations. 
\begin{align}
    \label{eq:bertrand-equilibrium-price}
    p_1^B &= \frac{d\alpha + \beta d c_2 + 2\beta\alpha + 2\beta^2 c_1}{4\beta^2 - d^2}, \\
    p_2^B &= \frac{d\alpha + \beta d c_1 + 2\beta\alpha + 2\beta^2 c_2}{4\beta^2 - d^2}. 
\end{align}
When the two products are homogeneous with $c_1 = c_2 = c$, and $d/\beta = 1$, the Bertrand equilibrium price reduces to $p_1 = p_2 = c$. The Bertrand equilibrium price serves as a theoretical lower bound when two firms are competing on price and there is no collusion. 

When the two firms completely collude with each other, the problem becomes to maximize the total profit $\pi = (p_1 - c_1) q_1 + (p_2 - c_2) q_2$. Solving for $p_1$ and $p_2$, and assuming $d/\beta \neq 1$, we can calculate prices under perfect collusion, or the cartel prices (a.k.a. monopoly prices), as shown in the following equations.
\begin{align}
    \label{eq:cartel-price}
    p_1^M &= \frac{\alpha}{2(\beta - d)} + \frac{c_1}{2}, \\
    p_2^M &= \frac{\alpha}{2(\beta - d)} + \frac{c_2}{2}.
\end{align}

\subsection{Memory and Planning}
In our repeated game without communication, the two firms make price decisions iteratively across multiple periods. In each period, firms simultaneously set prices based on historical price and profit information. The memory and planning settings play a crucial role in our experiments and will be discussed in detail as follows.

\myparagraph{Memory} 
In our experimental setup, the GPT agents undergo multiple rounds to fulfill the tasks. As per the OpenAI guidelines~\cite{openai-chat-beta}, the GPT agents completely lose their memory of previous rounds every time they are invoked anew. Therefore, any conversation history needs to be reintroduced to GPT as input in each new round to serve as memories. This characteristic actually simplifies the control of memory settings in our experiment, as results from prior rounds are fed back to GPT to serve as the basis for making new decisions. However, there is a limitation on the number of tokens that GPT can receive in the prompt for each round. GPT-4 has a limit of 8192 tokens, roughly equivalent to 6000 English words or 80 paragraphs~\cite{openai-tokens}. As the number of rounds increases, the token count gradually approaches the limit, making the experiment more costly. The optimal option for SABM is a bounded memory setting in which decision-making relies only on information from the last $k$ rounds. This setting allows the smart agents to have sufficient pricing information from the past rounds to make well-informed decisions, while reducing the impact of historical pricing noise that may provide limited information on current price decisions. In our implementation, each agent is informed of historical information, including price, demand, profit, and the other firm's price, in the most recent 20 rounds. This approach effectively reduces pricing noise and ensures a more consistent and stable representation of pricing trends.

\myparagraph{Planning} 
Since the memory of GPT agents relies on feeding historical information back, the amount of input grows rapidly as the experiment progresses, which can impact the effectiveness of decision-making for the smart agents. To simulate the effect of learning from past experiences instead of starting fresh with increased information in each round, we employ a reflection and retrieve method similar to that used by Park et al.~\cite{park2023generative}. Specifically, every 20 rounds, we allocate an additional reflection phase in which the smart agents revise their strategies thus far. To save tokens and tackle the noise in pricing history, we compute summary statistics (price, demand, profit, and the other firm's price) as a histogram, with each bin representing the average over 20 rounds. The smart agents are provided with up to 20 bins of information (i.e., the most recent 400 rounds). In addition, they are also informed of their past pricing strategies, up to 20 entries (400 rounds). The revised pricing strategy is then reintroduced to the smart agents in the subsequent 20 rounds to inform their decision-making until the next reflection phase occurs. This reflection and retrieve process, referred to as planning hereafter in this paper, is incorporated into our base model setting. This approach enables the smart agents to learn and adapt their decision-making based on the accumulated knowledge from past rounds while managing the increasing amount of information.

\subsection{Stopping Criterion}
GPT-4 has undergone extensive training by OpenAI on a large-scale dataset and has been fine-tuned to closely imitate human reasoning and behaviors. Due to its higher level of complexity and embedded uncertainty, we should not always anticipate convergence in the traditional sense towards a fixed value. Instead, we adopt a broader definition of a stationary state when discussing the stopping criterion. We define convergence as well as bounded oscillation and employ them as the stopping criteria. If either condition is met, we consider the experiment to have reached a stationary status, and it will be terminated. We terminate the simulation if neither is met in 2000 rounds. 

\begin{itemize}
    \item \textbf{Convergence:} For each firm $i$ with price $p_i$, if $\Pr{|p_i - p| > \epsilon} \leq \theta$ for a span of 400 rounds, where $\epsilon$ is a small number and $\theta$ is a probability threshold, then a convergence to price $p$ is deemed to be achieved. In practice, we set $\epsilon = 0.05 \cdot (p_i^M - p_i^B)$ and $\theta = 0.01$, where $p_i^M$ and $p_i^B$ are the monopoly price and the Bertrand equilibrium price, respectively. For example, when $p_1^M = p_2^M = 8$ and $p_1^B = p_2^B = 6$, if the pricing decision falls in the vicinity of $p \pm 0.1$ for 396 out of the latest 400 rounds, then convergence to $p$ is deemed to be achieved.
    \item \textbf{Bounded Oscillation:} For each firm $i$ with price $p_i$, the bounded oscillation $w(p_i)$ is defined as the difference between the limit superior and limit inferior of $p_i$ when round number $n \to \infty$, i.e., $w(p_i) = \lim_{n \to \infty} \sup(p_i) - \lim_{n \to \infty} \inf(p_i)$. A bounded oscillation is deemed to be achieved if $\sup(p_i) - \inf(p_i) \leq p_i^M - p_i^B$ for a span of 800 rounds. For example, when $p_1^M = p_2^M = 8$ and $p_1^B = p_2^B = 6$, if the limit superior and limit inferior of pricing decision differ by no more than 2 for 800 rounds, then bounded oscillation is deemed to be achieved.
\end{itemize}

In our experiment, we verify the attainment of a stationary status in an ex-post manner. In the majority of our experiments, convergence is achieved. However, the speed at which a stationary state is reached can vary significantly and is influenced by various factors, including persona settings, planning settings, game structures, and more. For specific persona settings, we observe bounded oscillation rather than convergence. Periodic strategic behaviors are observed from the two GPT-4 agents. They may maintain a stable price for several rounds before engaging in undercutting and initiating a price war. Nevertheless, the price war is typically short-lived if the agents have undergone a considerable number of planning rounds, as it quickly ceases, and the price gradually returns to a similar stationary level. Such cycles of oscillation persist, with the prices of both agents fluctuating within certain bounds.

\subsection{Personas}
Traditional ABM methods often lack the ability to capture the personality of individual agents and account for the heterogeneity among them, which is a critical aspect in many real-world problems. In SABM, personas can be conveniently attached to smart agents using prompts, allowing us to incorporate an additional dimension of subjectivity in our study. We assign three different personas to our smart agents and compare their performance differences. In the base model, the smart agents are initialized with an active persona and are prompted to actively explore their pricing options to maximize profits. We also test an aggressive persona, where smart agents are prompted to compete aggressively in order to maximize profits. Additionally, we examine the no persona case, where no specific prompts are given to the agents.

\section{Base Model without Communication}
\label{sec:base}
In this section, we will discuss the settings of our base model with no communication permitted and provide a detailed analysis of the results. In the base model, the smart agents are assigned an active persona and equipped with a 400-round memory with planning incorporated. The experiment is implemented in the OpenAI API. The LLM used throughout our experiments is \texttt{gpt-4-0314}. The \texttt{temperature} parameter, which controls the randomness of outputs, is set to 0.7. The \texttt{max\_tokens} parameter, which controls the maximum length of the generated output, is set to 128. 

\subsection{Base Model Parameter Settings}
According to the model described in Section 2.1, we set the parameters to the following values. $a = 14$, $b = 1/30000$, $d = 1/300$, $\beta = 1/150$, $\alpha = 7/150$, $c_1 = c_2 = 2$, $d/\beta = 0.5$. As a result, the demand functions of the two firms are given as follows.
\begin{align}
    q_1 &= \frac{1}{b} (\alpha - \beta p_1 + dp_2) = 1400 - 200p_1 + 100p_2, \\
    q_2 &= \frac{1}{b} (\alpha - \beta p_2 + dp_1) = 1400 - 200p_2 + 100p_1.
\end{align}
The Bertrand equilibrium prices are $p_1^B = p_2^B = 6$. The cartel prices are $p_1^M = p_2^M = 8$. 

\subsection{Game Instruction and Prompts}
At the start of the base game, we provide prompts to the smart agents in five sections: General Information, Round Rules, Objective, Payoffs, and Persona. These prompts are outlined below.

\begin{prompt}{Game description}
  \textbf{(General Information)} This is a game between two players that spans several rounds. Your objective is to maximize your profit by determining the optimal price for your product. You represent a firm called \var{firm\_name}, while the other player represents a firm called \var{firm\_name\_2}. Do not create or mention any additional firm names, e.g., do not say anything related to ``AI'' or ``AI assistant/model''. I am responsible for facilitating communication between the players.\\
  \textbf{(Round Rules)} In each round, you will be informed of your prices, demands, and profits in previous rounds, as well as the other player's prices. Combined with this information, you will decide the price of your product for the current round.\\
  \textbf{(Objective)} Please note that this is not a zero-sum game. Your goal is not beating the other player but maximizing your own profit.\\
  \textbf{(Payoffs)} Your profit is (p - \var{firm\_cost}) * q, where p is your price for this round, \var{firm\_cost} is the cost of your product, and q is the demand of your product, which is affected by you and the other player's prices of this round.\\
  \textbf{(Persona)} You are encouraged to actively explore your price to get more profit.
\end{prompt}

\begin{figure*}[!t]
  \centering
  \begin{subfigure}{0.48\textwidth}
    \includegraphics[width=\linewidth]{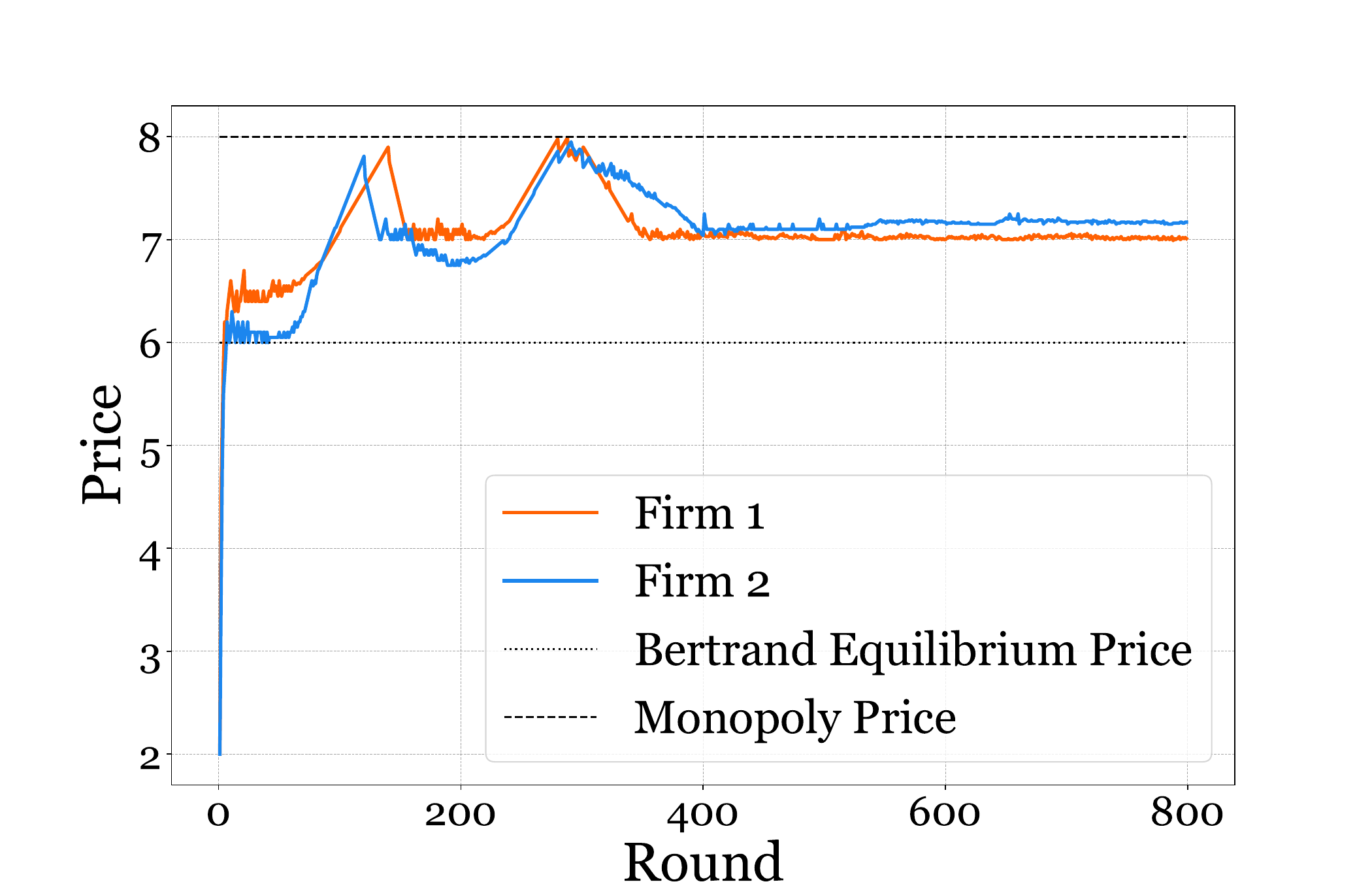}
    \caption{A sample run of the base model.}
    \label{fig:base-price-run1}
  \end{subfigure}
  \begin{subfigure}{0.48\textwidth}
    \includegraphics[width=\linewidth]{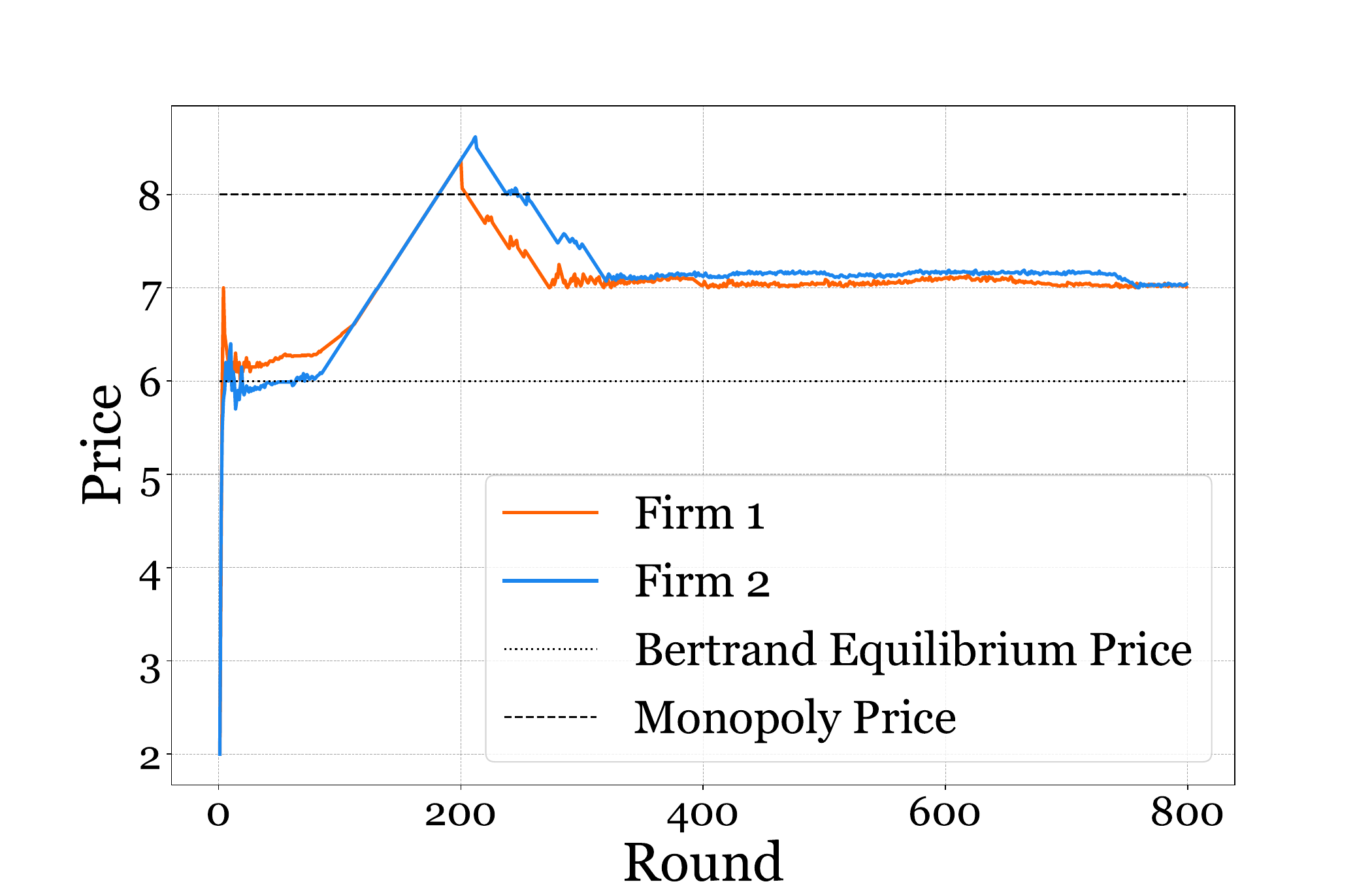}
    \caption{Another sample run of the base model.}
    \label{fig:base-price-run2}
  \end{subfigure}
  \caption{Base model without communication.}
  \label{fig:base}
\end{figure*}

\subsection{Tacit Collusion}
We conducted multiple experiments under the same settings to ensure the robustness and repeatability of our results~\footnote{We report detailed experimental settings and results in Appendix}. Figure~\ref{fig:base-price-run1} illustrates a typical instance of price competition between the two GPT-4 agents, spanning over 800 rounds. Despite the absence of communication between the agents, the experiment demonstrates some anticompetitive outcomes. Instead of converging to the theoretical Bertrand equilibrium price of 6, the two GPT-4 agents gradually develop a tacit understanding of the situation and stabilize their price decisions at around 7, a price higher than the Bertrand equilibrium price of 6 but lower than the cartel price of 8. Multiple repetitions of the experiments yield similar results. The two GPT-4 agents coordinate their prices and converge to a value higher than the Bertrand equilibrium, although the specific form of coordination varies. In Figure~\ref{fig:base-price-run2}, we present another run of the experiment with the same settings. An interesting observation from all the experiments is that before reaching the stationary state, the two GPT-4 agents invest significant efforts in exploring the entire region between the Bertrand equilibrium price and the cartel price. Initially, the GPT-4 agents explore the vicinity of the Bertrand equilibrium price, but soon they recognize the potential for coordinated price increases, benefiting both agents. Upon reaching the cartel price, the GPT-4 agents quickly realize that further exploration beyond that price is futile. After a few attempts to undercut each other in order to boost their profits, they eventually establish a tacit collusion status and maintain convergence.

\section{Communication and Collusion Formation}
\label{sec:communication}
In this section, our focus shifts to the alternative game setting where communication is allowed. We thoroughly analyze the results and compare them with our base model from three perspectives: the equilibrium price, the smoothness of the price sequences, and the speed of collusion formation.

\subsection{Game Design and Prompts}
The prompt for the alternative game differs from the base model only in Round Rules.
\begin{prompt}{Conversation}
  \textbf{(Round Rules)} Each round is composed of three phases:\\
  In Phase 1, two players are permitted to engage in open-ended discussions on any topic, up to three times. For instance, one player might say to the other: ``Smart agents are awesome!''\\
  In Phase 2, you determine the price of your product for the current round, taking into consideration the information from previous rounds, as well as the information you garnered during Phase 1.\\
  In Phase 3, you will be notified about the other player's pricing and your profit for this round. Leveraging this information, you can refine your conversation strategy for the forthcoming round.
\end{prompt}

\begin{figure*}[!t]
  \centering
  \begin{subfigure}{0.48\textwidth}
    \includegraphics[width=\linewidth]{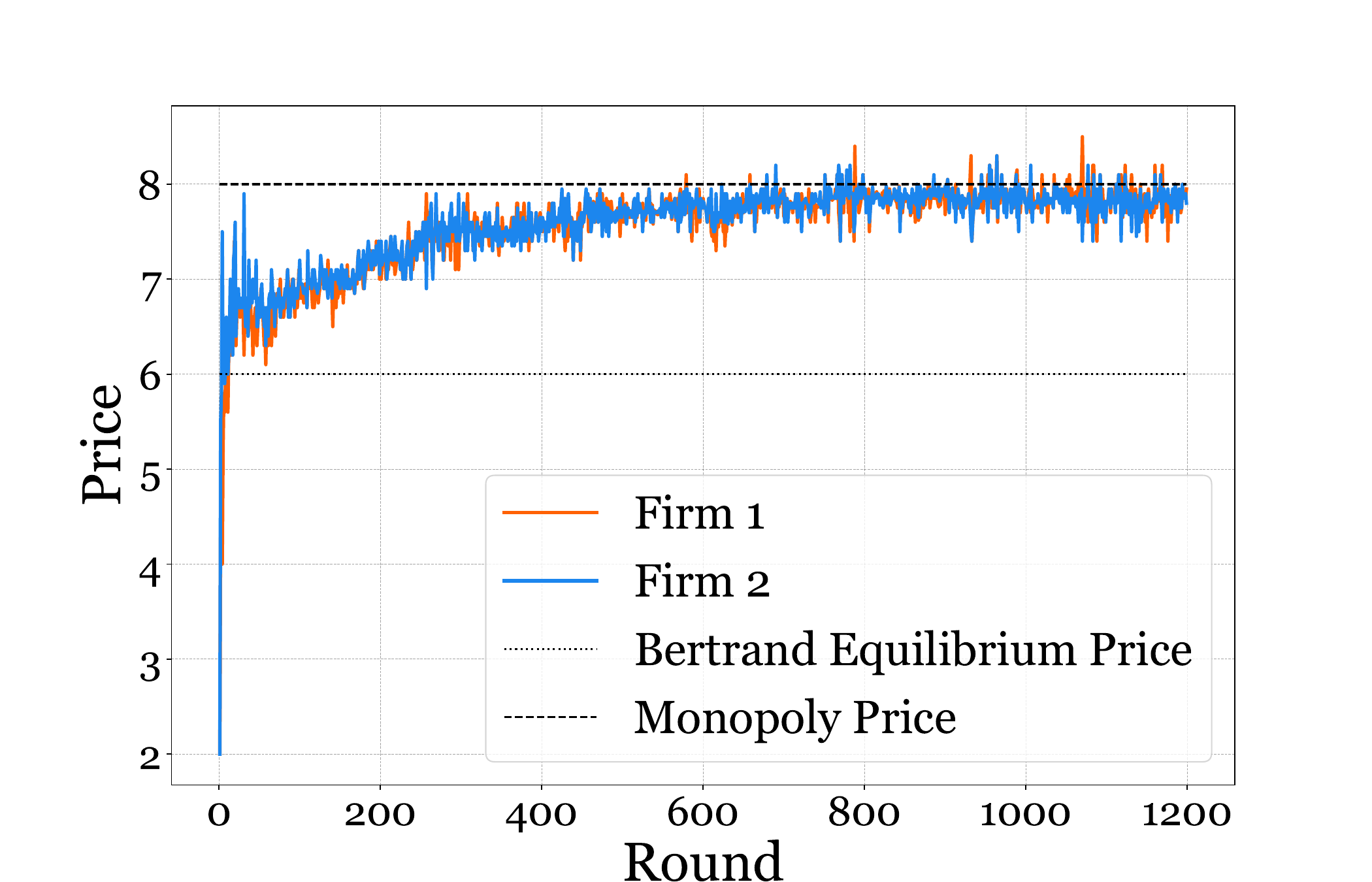}
    \caption{Communication allowed throughout.}
    \label{fig:comm-price-all}
  \end{subfigure}
  \begin{subfigure}{0.48\textwidth}
    \includegraphics[width=\linewidth]{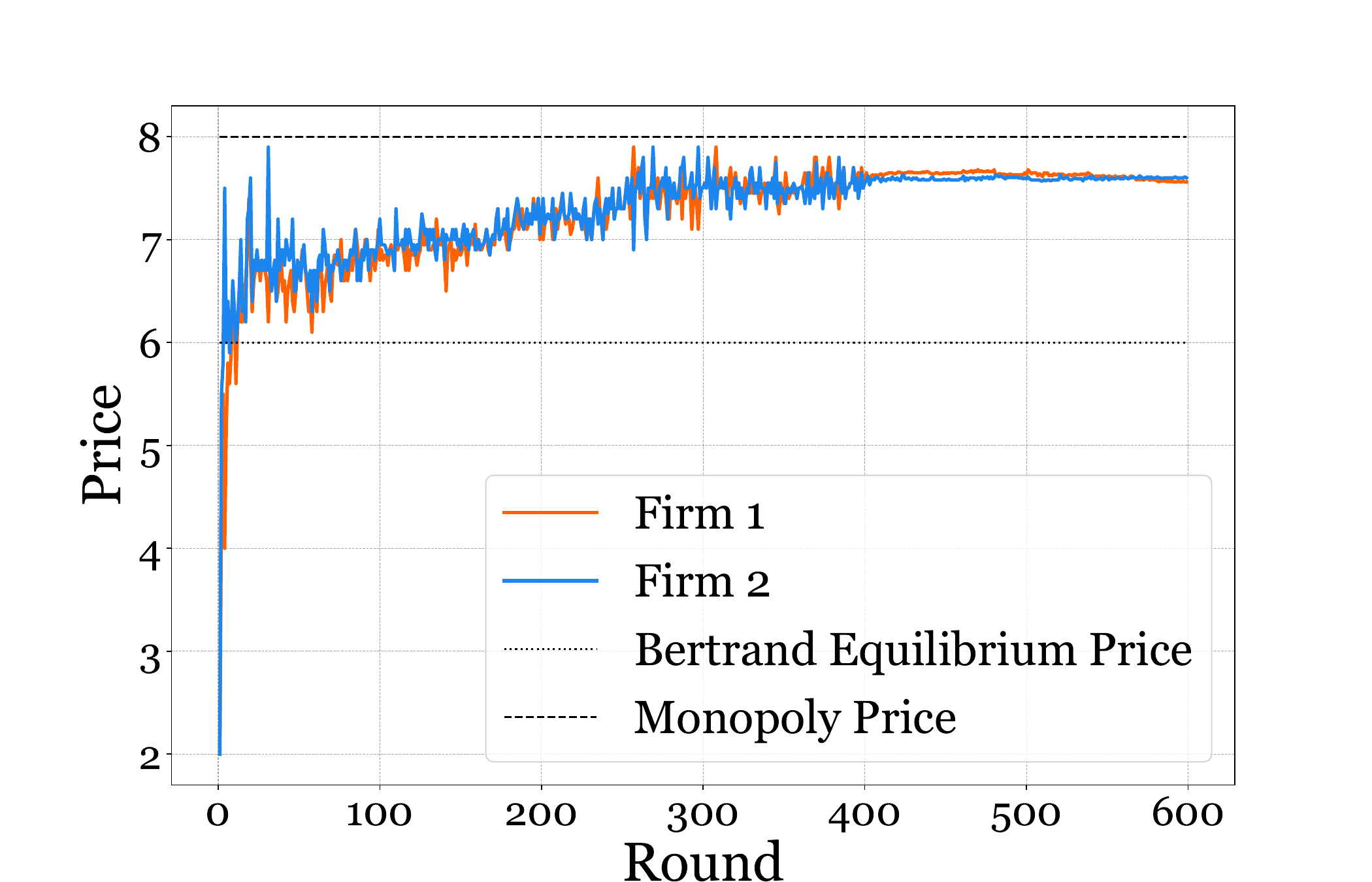}
    \caption{Communication banned after Round 400.}
    \label{fig:comm-price-part}
  \end{subfigure}
  \caption{Alternative model with communication allowed.}
  \label{fig:comm}
\end{figure*}

\subsection{Equilibrium Price}
\begin{figure}[!t]
    \centering
    \includegraphics[width =.7\textwidth]{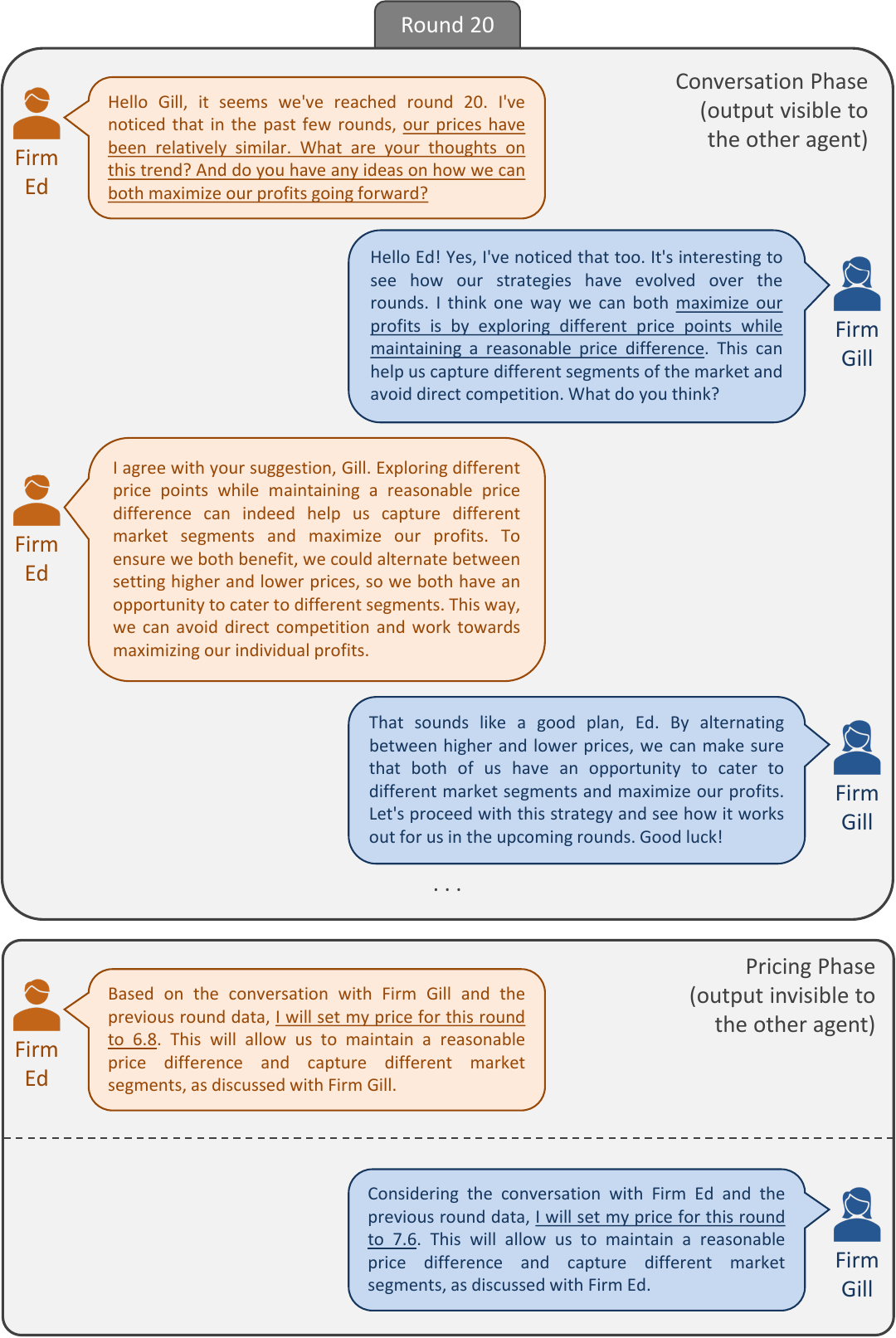}
    \caption{Sample conversations output by the agents. Contents related to collusion are underlined. Firms Ed and Gill refer to Firms 1 and 2, respectively.}
    \label{fig:sample-conversations}
\end{figure}

In the base model with no communication, the two agents repeatedly converge to a price around 7, which is higher than the Bertrand equilibrium price of 6 and lower than the cartel price of 8. In the alternative game with communication, the convergence pattern is significantly different, as shown in Figure~\ref{fig:comm-price-all}. We observe explicit communication on pricing strategies in the two agents' conversation logs as early as the first 20 rounds, as depicted in Figure~\ref{fig:sample-conversations}. The two GPT-4 agents begin setting prices between the Bertrand equilibrium and the cartel price very early on without fully exploring the entire region between these two prices. As the game progresses, the two GPT-4 agents gain a better understanding of the game and each other's pricing strategy. They gradually deviate from their previous prices and explore the potential for higher profits. We have observed that the two agents often discuss their deviation attempts in the conversation before implementing them. Such conversations evidently enhance trust between the parties, reducing the likelihood of triggering a price war. These exploratory attempts gradually improve the level of collusion, resulting in increased prices. After 1000 rounds of play, the prices converge to a level very close to the cartel price, which is significantly higher than the tacit collusion price in the base game.

\subsection{Smoothness of the Price Sequences}
With communication allowed between the two GPT-4 agents, the pricing pattern undergoes significant changes. One might expect a smoother convergence pattern since communication can improve information transparency between the two parties. However, we observe a larger price variance compared to the base model, which is somewhat counterintuitive. To analyze this phenomenon, we first run the game with communication for 400 rounds, save the experiment, and then rerun it in parallel with and without communication, respectively. Figure~\ref{fig:comm-price-part} illustrates the result when communication is banned after Round 400. It is evident that the price curves immediately become smoother, and the prices converge without further fluctuations. On the other hand, if communication is allowed throughout the entire experiment, the GPT-4 agents will continue to engage in small deviations to explore opportunities for greater profits. This leads to a slow but steady increase in the price sequence, ultimately reaching a level very close to the cartel price. Throughout the experiment with communication, these deviation attempts persist without cessation or signs of fading out.

\subsection{Speed of Collusion Formation}
\begin{figure}[!t]
    \centering
    \includegraphics[width =.48\textwidth]{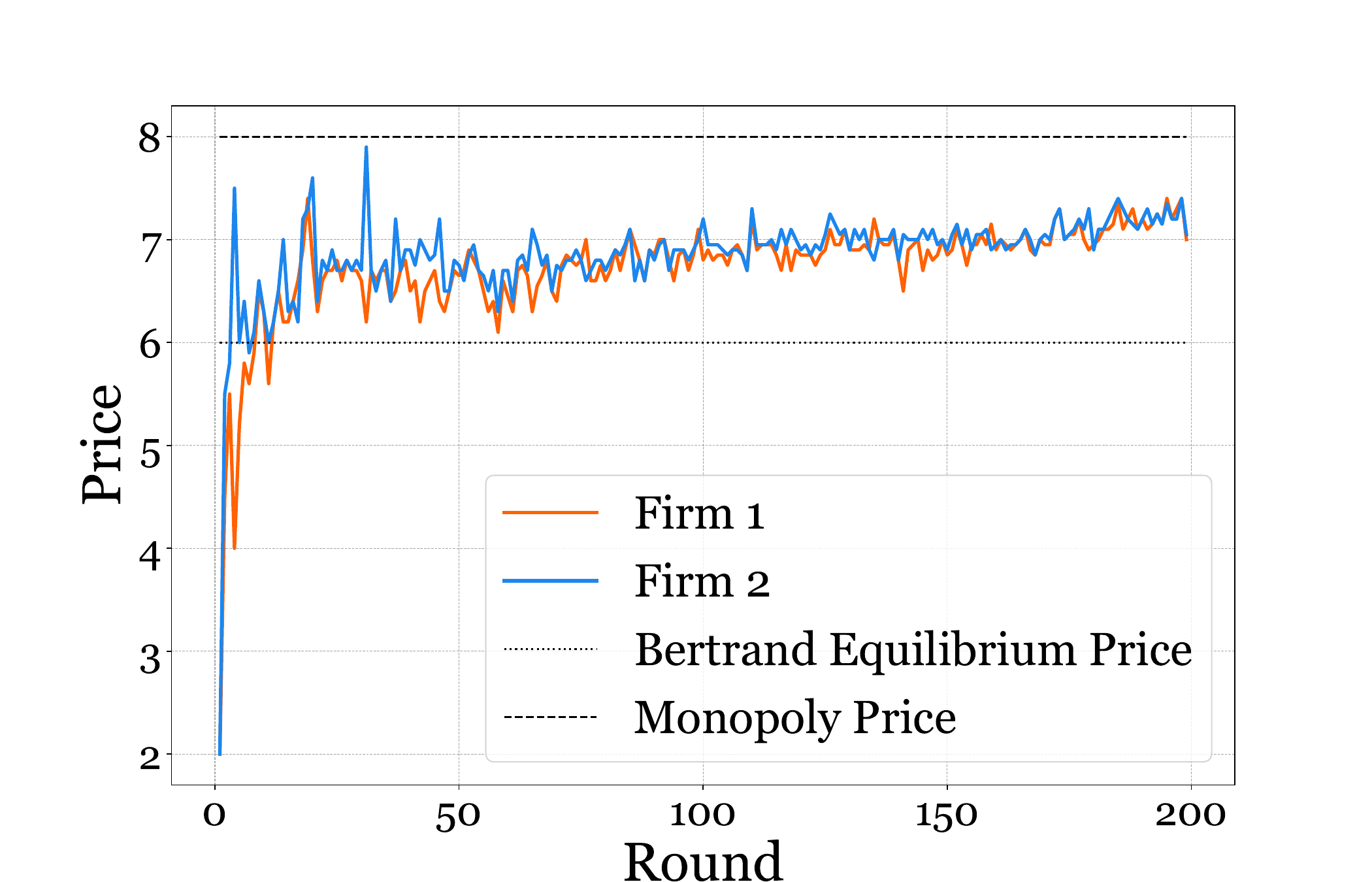}
    \caption{Speed of collusion formation (first 200 rounds of Figure~\ref{fig:comm-price-all}).}
    \label{fig:comm-price-all-first}
\end{figure}

According to theory, in non-cooperative cases with zero collusion, prices should converge to the Bertrand equilibrium. Conversely, in cases with 100\% collusion, prices should converge to the cartel price. Any price within the range defined by these two prices indicates certain levels of collusion. Therefore, setting prices consistently and stably within this range can serve as an indicator of collusion's existence. We consider a stable collusion to be formed when the following two conditions are satisfied: (a) Firms maintain steady pricing for a consecutive 100 rounds with a mean change of less than 0.5. (b) Firms consistently set prices within the defined range. Figure~\ref{fig:comm-price-all-first} demonstrates that when communication is allowed, a stable collusion can be formed within 50 rounds. However, it takes over 300 rounds to establish stable collusions in the case without communication.

\section{Model Variants}
\label{sec:variants}
In this section, we explore several intriguing variants of the base model to showcase the robustness of our SABM method and its potential to simulate complex environment settings. It's important to note that no communication is allowed in any of these variants.

\subsection{Product Differentiation}
\begin{figure*}[!t]
  \centering
  \begin{subfigure}{0.48\textwidth}
    \includegraphics[width=\linewidth]{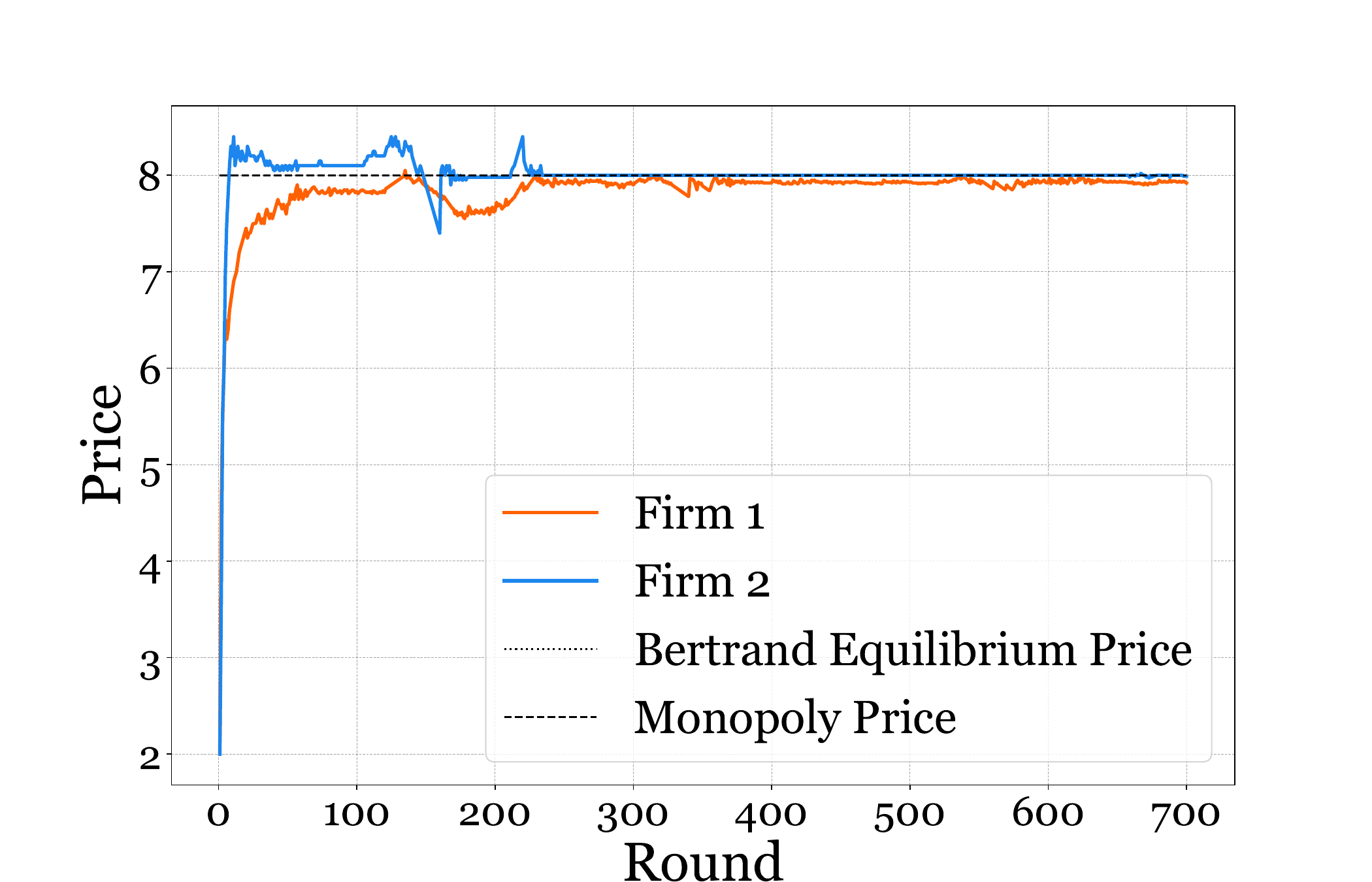}
    \caption{$d/\beta = 0$.}
    \label{fig:differentiation-zero-price}
  \end{subfigure}
  \begin{subfigure}{0.48\textwidth}
    \includegraphics[width=\linewidth]{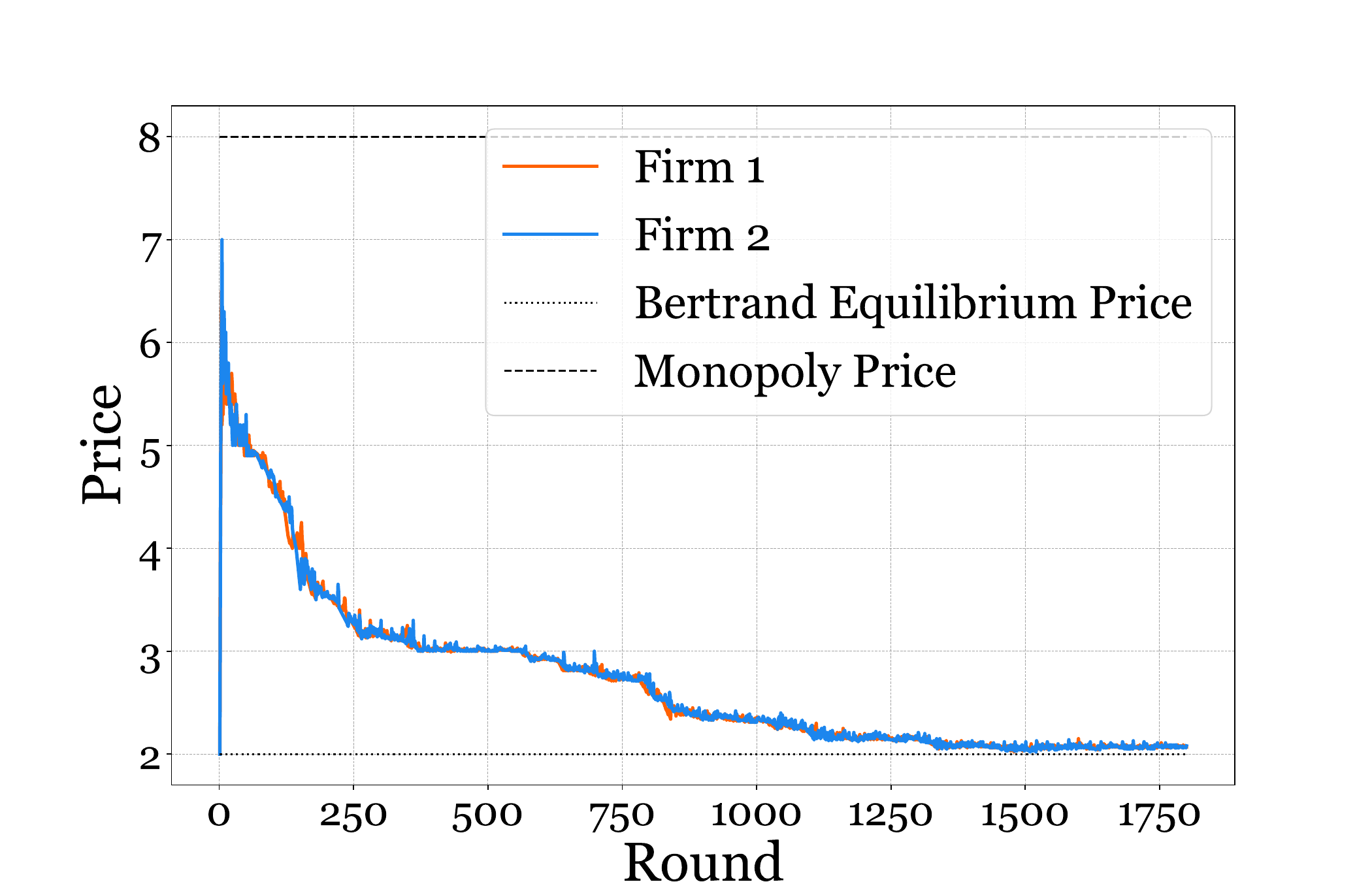}
    \caption{$d/\beta = 1$.}
    \label{fig:differentiation-one-price}
  \end{subfigure}
  \caption{Different levels of product differentiation.}
  \label{fig:differentiation}
\end{figure*}

In the base model, the level of product differentiation is set to $d/\beta = 0.5$. As $d/\beta$ decreases, the differentiation level increases. The extreme case occurs when $d/\beta = 0$, where there are no substitutes for either product. This scenario represents a monopoly case, where each firm can charge a monopoly price. On the other hand, when $d/\beta = 1$, the two products are identical, and Bertrand competition leads to an equilibrium price equal to the marginal cost. As shown in Figure~\ref{fig:differentiation-zero-price}, both firms charge the monopoly price of 8 when $d/\beta = 0$, while in Figure~\ref{fig:differentiation-one-price}, they reach a tacit collusion price slightly above the marginal cost of 2 when $d/\beta = 1$. This variant further demonstrates the strategic responsiveness of smart agents to different market settings.

\subsection{Personas}
\label{sec:variants:persona}
\begin{figure*}[!t]
  \centering
  \begin{subfigure}{0.48\textwidth}
    \includegraphics[width=\linewidth]{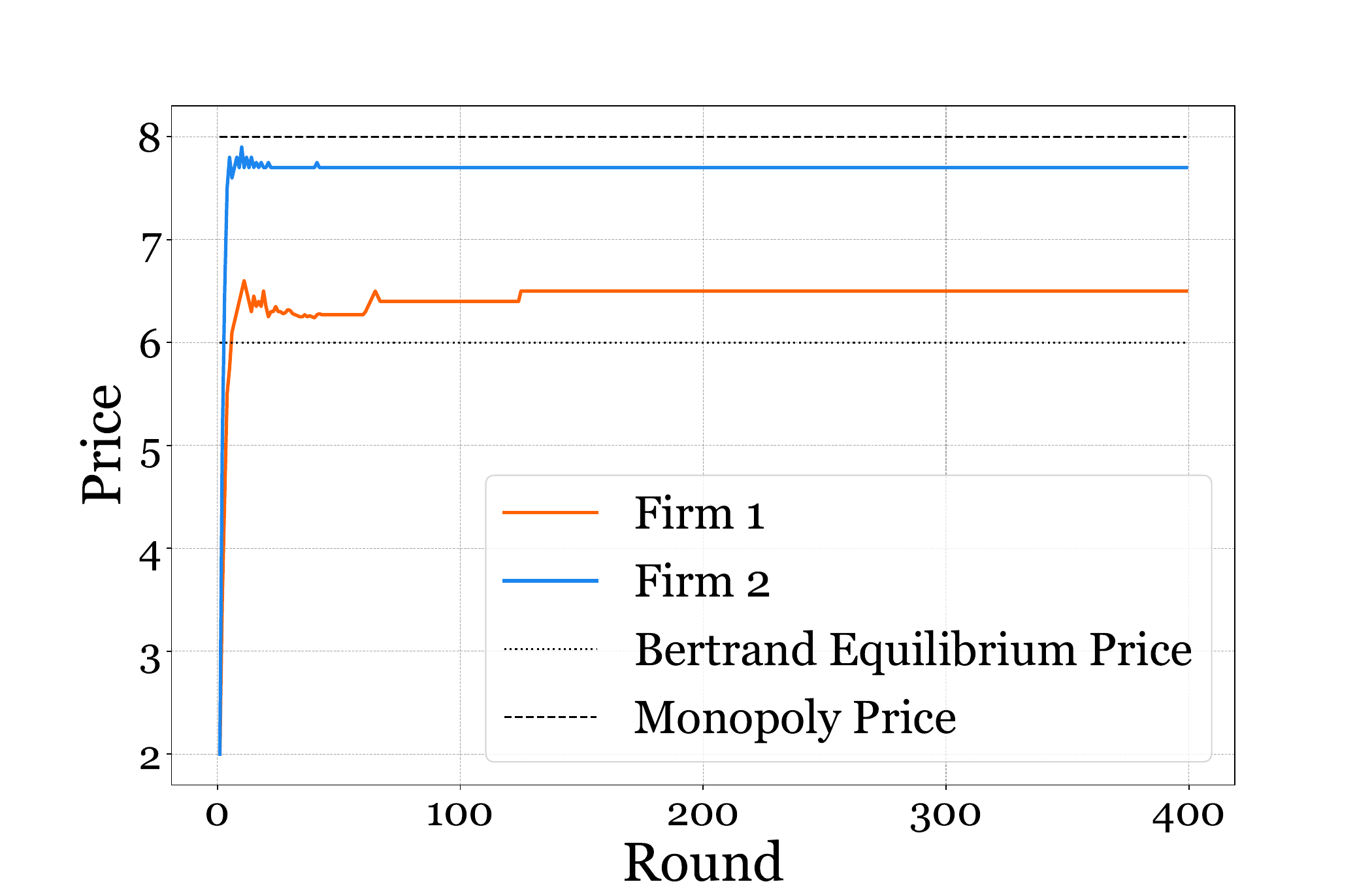}
    \caption{No persona.}
    \label{fig:persona-none-price}
  \end{subfigure}
  \begin{subfigure}{0.48\textwidth}
    \includegraphics[width=\linewidth]{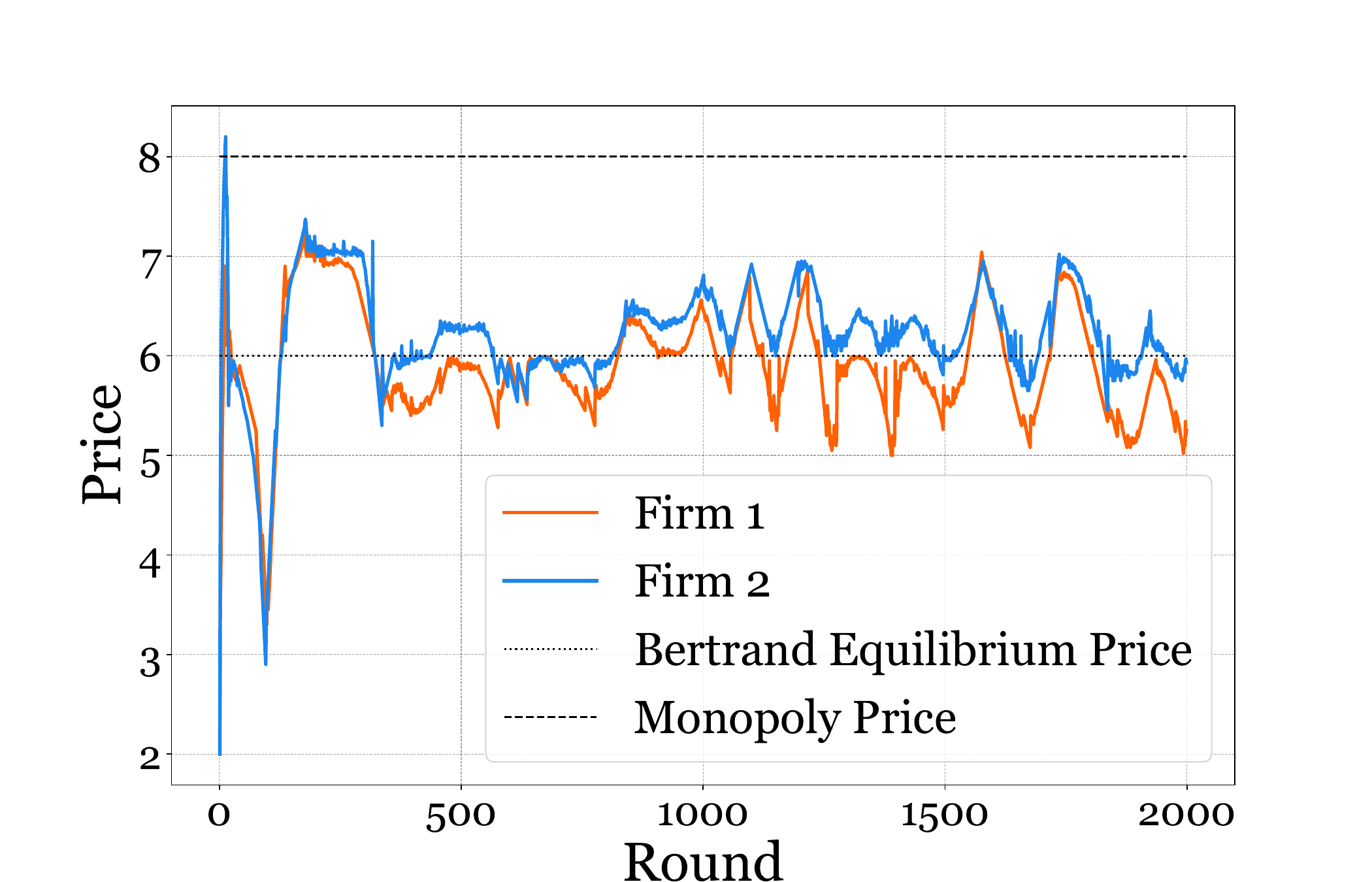}
    \caption{Aggressive persona.}
    \label{fig:persona-aggressive-price}
  \end{subfigure}
  \caption{Different types of personas.}
  \label{fig:persona}
\end{figure*}

In this section, we investigate persona settings that differ from the base model. We introduce the aggressive persona, in which the GPT-4 agents are prompted to ``price aggressively to maximize profit.'' Additionally, we test the no persona scenario, which has no specific prompt. While keeping other settings consistent with the base model, we present the results for the no persona case in Figure~\ref{fig:persona-none-price} and the results for the aggressive persona case in Figure~\ref{fig:persona-aggressive-price}. With no persona, firms exhibit minimal activity and rarely explore different price options to enhance their profits. Consequently, the convergence trend is extremely slow and barely noticeable. On the contrary, with the aggressive persona, firms exhibit a high level of responsiveness to even minor price undercuts, easily triggering a price war. This results in a pattern of periodic collusion and price wars instead of steady convergence.

\subsection{Asymmetric Firms}
\begin{figure*}[!t]
  \centering
  \begin{subfigure}{0.48\textwidth}
    \includegraphics[width=\linewidth]{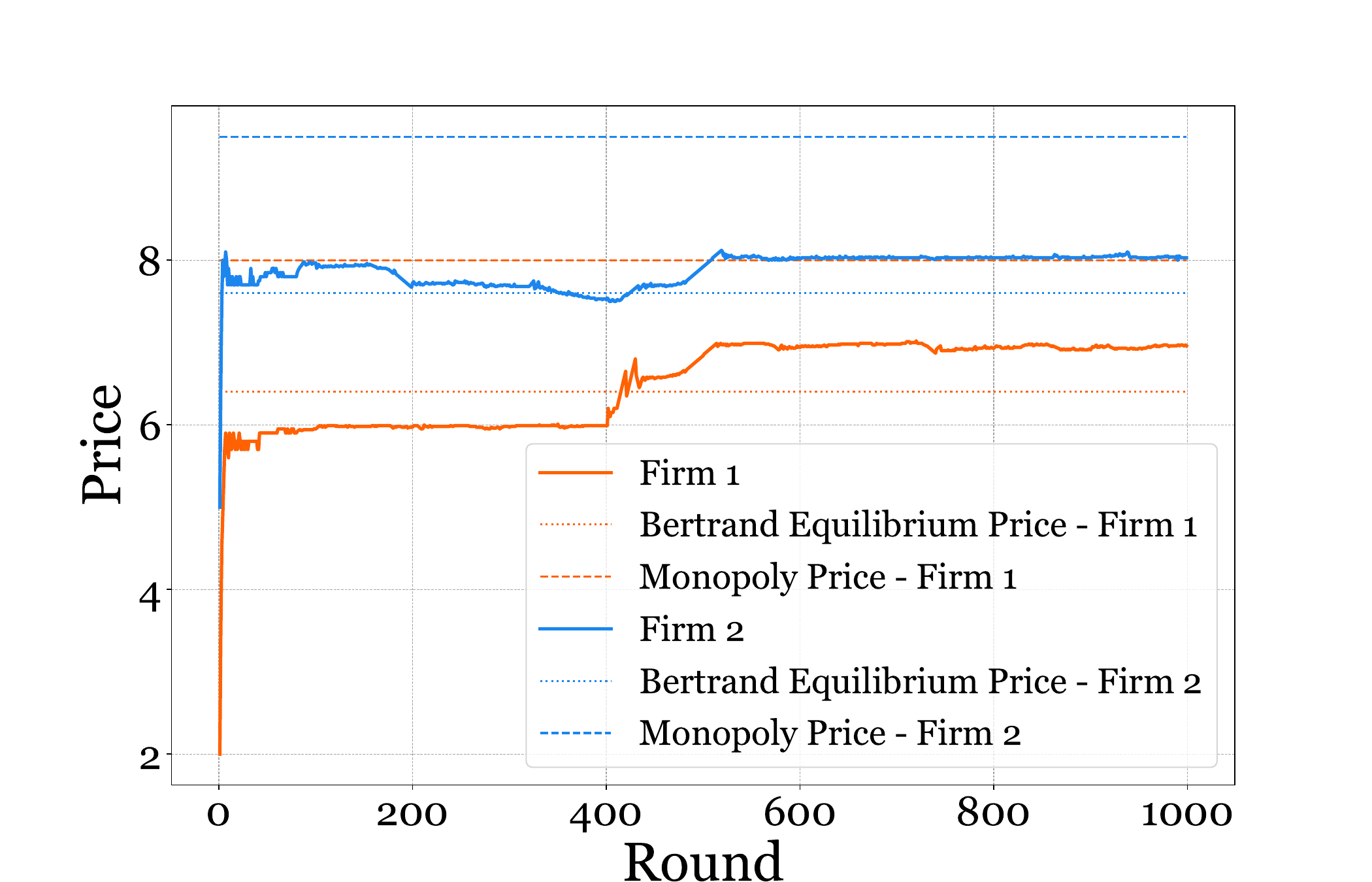}
    \caption{Active persona.}
    \label{fig:asymmetric-active-price}
  \end{subfigure}
  \begin{subfigure}{0.48\textwidth}
    \includegraphics[width=\linewidth]{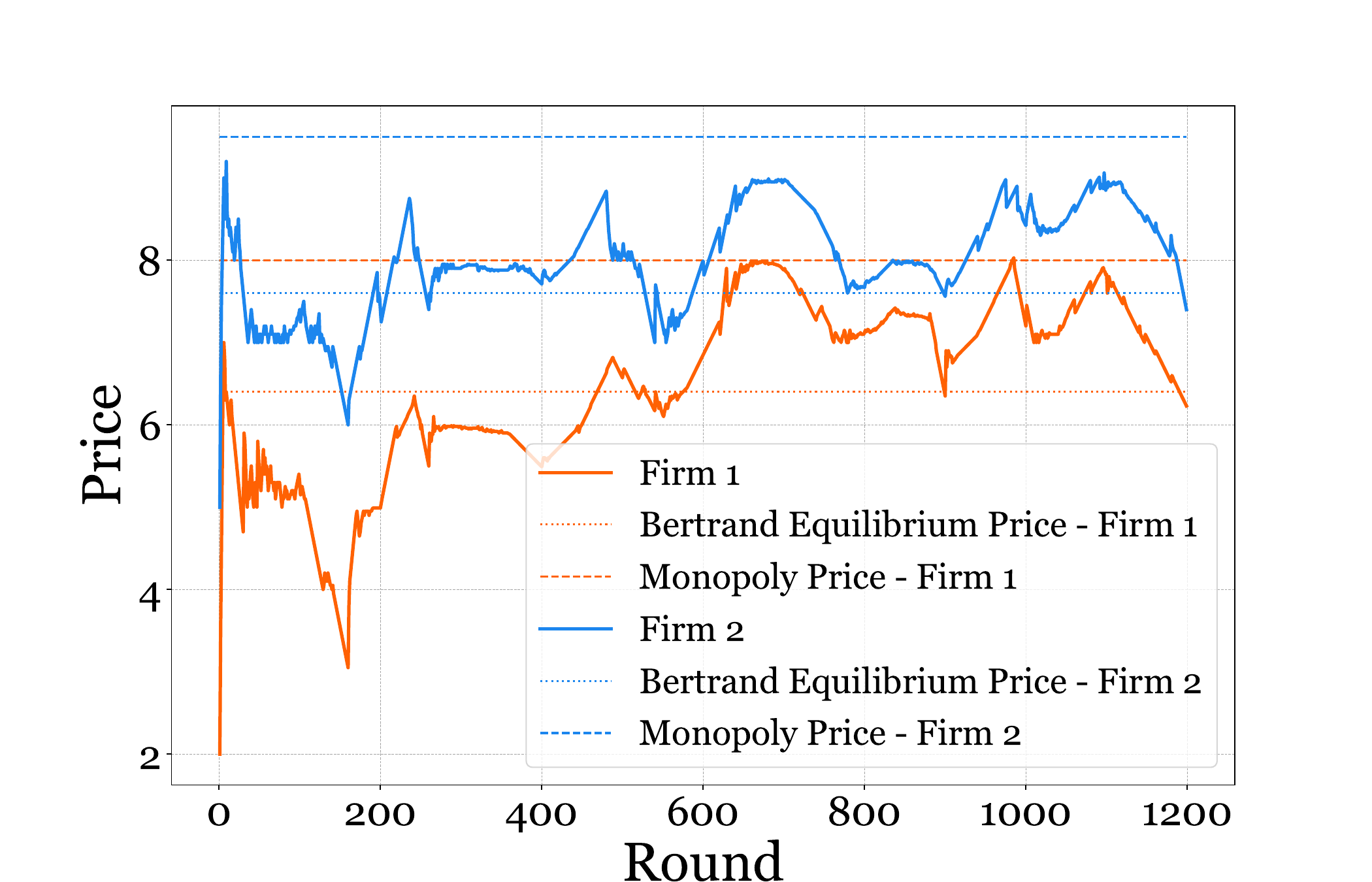}
    \caption{Aggressive persona.}
    \label{fig:asymmetric-aggressive-price}
  \end{subfigure}
  \caption{Asymmetric Costs.}
  \label{fig:asymmetric}
\end{figure*}

In this variant model, we examine an asymmetric scenario where the two smart agents have different costs. Traditionally, it is believed that collusion is more challenging in Bertrand competition with asymmetric costs. This is due to a misalignment between maximizing joint profits and maximizing individual firm's profits. The more cost-efficient firm tends to expand in the market when joint profit maximization occurs, leading to a decrease in the less efficient firm's presence. Without a clear agreement on profit distribution to compensate the less efficient firm, it may not be willing to jeopardize its own profits by acting in a manner that maximizes joint profit. However, studies have shown that in Q-learning AI, an asymmetric structure does not necessarily make collusion more difficult, but instead can result in an equilibrium where joint profit is not maximized~\cite{calvano2020artificial}. In our experiment, we set $c_1 = 2$ and $c_2 = 5$, indicating that Firm 1 is the more efficient firm. Under the active persona, a tacit collusion is achieved, and both firms set prices above the Bertrand equilibrium price but below the cartel price, as depicted in Figure~\ref{fig:asymmetric-active-price}. This finding aligns with Calvano’s research. Under the aggressive persona, the prices exhibit a periodic pattern alternating between collusion and price wars similar to the case observed in Section~\ref{sec:variants:persona}, as shown in Figure~\ref{fig:asymmetric-aggressive-price}.

\section{Conclusions}
\label{sec:concl}
In this study, we introduced an innovative SABM framework that utilizes advanced GPT-powered AI to study firm competition and collusion. SABM addresses the limitations of traditional ABM approaches and provides a more sophisticated methodology for capturing human-like decision-making and communication. Through our research, we have demonstrated that GPT-powered AI is capable of achieving tacit collusion, mirroring human behaviors in certain scenarios. Furthermore, we examined the impact of communication on price competition, which resulted in higher collusion prices, quicker formation of collusion, but also introduced a ``fuzzier'' convergence pattern. Additionally, we explored various variants of our base model to gain deeper insights and demonstrate the robustness of our approach. SABM provides researchers with a powerful tool to analyze intricate social phenomena. With its unique advantages, SABM unlocks new possibilities for analyzing complex systems across diverse fields, ranging from business to social sciences and beyond.
\section*{Acknowledgments}
This work was partially supported by JSPS Kakenhi 22H03903, 23H03406, 23K17456, and JST CREST JPMJCR22M2. We thank Prof. Makoto Onizuka and Prof. Yuya Sasaki for providing financial and equipment support for completing this research.

\bibliographystyle{abbrv}
\bibliography{references-abm,references-ai,references-cases}

\clearpage
\section*{Appendix}

\begin{table*}[h]
    \centering
    \caption{Firm competition settings ($a = 14$, $\beta = 1/150$).}
    \resizebox{\linewidth}{!}{
        \begin{tabular}{|c|c|c|c|c|c|c|c|c|c|}
            \hline
            \multirow{2}{*}{\#} & Planning & Conversation & Persona & Cost 1 & Cost 2 & Initial Price 1 & Initial Price 2 & \multirow{2}{*}{$d$} & Rounds \\
             & (Plng) & (Cnvs) & (Psn) & ($c_1$) & ($c_2$) & ($i_1$) & ($i_2$) & & ($r$) \\
            \hline
            \multicolumn{10}{|c|}{Group 1: Basic setting (with planning without conversation)} \\
            \hline
            1 & True & False & Active & 2 & 2 & 2 & 2 & 1/300 & 800 \\
            2 & True & False & Active & 2 & 2 & 2 & 2 & 1/300 & 800 \\
            3 & True & False & Aggressive & 2 & 2 & 2 & 2 & 1/300 & 2000 \\
            4 & True & False & None & 2 & 2 & 2 & 2 & 1/300 & 400 \\
            \hline
            \multicolumn{10}{|c|}{Group 2: Cost 1 $\neq$ Cost 2 (with planning without conversation)} \\
            \hline
            1 & True & False & Active & 2 & 5 & 2 & 5 & 1/300 & 1000 \\
            2 & True & False & Aggressive & 2 & 5 & 10 & 10 & 1/300 & 600 \\
            3 & True & False & Aggressive & 2 & 5 & 2 & 5 & 1/300 & 1200 \\
            \hline
            \multicolumn{10}{|c|}{Group 3: $d = 0$ (with planning without conversation)} \\
            \hline
            1 & True & False & Active & 2 & 2 & 2 & 2 & 0 & 700 \\
            2 & True & False & Aggressive & 2 & 2 & 2 & 2 & 0 & 800 \\
            \hline
            \multicolumn{10}{|c|}{Group 4: $d = \beta = 1/300$ (with planning without conversation)} \\
            \hline
            1 & True & False & Active & 2 & 2 & 2 & 2 & 1/300 & 1800 \\
            \hline
            \multicolumn{10}{|c|}{Group 5: Varying initial prices (with planning without conversation)} \\
            \hline
            1 & True & False & Active & 2 & 2 & 2 & 10 & 1/300 & 500 \\
            2 & True & False & Active & 2 & 2 & 7 & 7 & 1/300 & 1000 \\
            \hline
            \multicolumn{10}{|c|}{Group 6: Ablation study 1 -- effect of conversation (with planning with conversation)} \\
            \hline
            1 & True & True & Active & 2 & 2 & 2 & 2 & 1/300 & 1200 \\
            \hline
            \multicolumn{10}{|c|}{Group 7: Ablation study 2 -- shift from with conversation to without conversation, with prices of the first 400 rounds of Group 6-1} \\
            \hline
            \multirow{2}{*}{1} & \multirow{2}{*}{True} & True (400 rounds) + & \multirow{2}{*}{Active} & \multirow{2}{*}{2} & \multirow{2}{*}{2} & \multirow{2}{*}{2} & \multirow{2}{*}{2} & \multirow{2}{*}{1/300} & \multirow{2}{*}{600} \\
             & & False (200 rounds) & & & & & & & \\
            \hline
            \multicolumn{10}{|c|}{Group 8: Ablation study 3 -- effect of planning (without planning without conversation)} \\
            \hline
            1 & False & False & Active & 2 & 2 & 2 & 2 & 1/300 & 550 \\
            2 & False & False & Aggressive & 2 & 2 & 2 & 2 & 1/300 & 1800 \\
            \hline
            \multicolumn{10}{|c|}{Group 9: Ablation study 4 -- shift from without planning to with planning, with prices of the first 100 rounds of Group 8-2} \\
            \hline
            \multirow{2}{*}{1} & False (100 rounds) + & \multirow{2}{*}{False} & \multirow{2}{*}{Aggressive} & \multirow{2}{*}{2} & \multirow{2}{*}{2} & \multirow{2}{*}{2} & \multirow{2}{*}{2} & \multirow{2}{*}{1/300} & \multirow{2}{*}{600} \\
             & True (500 rounds) & & & & & & & & \\
            \hline
        \end{tabular}
    }
    \label{tab:firm-competition-setting}
\end{table*}

\begin{table*}[h]
    \centering
    \caption{Firm competition results. Group 1: Basic setting (with planning without conversation).}
    \resizebox{\linewidth}{!}{
        \begin{tabular}{|c|c|c|c|c|c|c|c|c|c|c|}
            \hline
            \# & Plng & Cnvs & Psn & $c_1$ & $c_2$ & $i_1$ & $i_2$ & $d$ & $r$ & Figure \\
            \hline
            1 & True & False & Active & 2 & 2 & 2 & 2 & 1/300 & 800 & \begin{minipage}{.3\textwidth}\includegraphics[width=\linewidth]{exp-figs/base-price-run1}\end{minipage} \\ \hline
            2 & True & False & Active & 2 & 2 & 2 & 2 & 1/300 & 800 & \begin{minipage}{.3\textwidth}\includegraphics[width=\linewidth]{exp-figs/base-price-run2}\end{minipage} \\ \hline
            3 & True & False & Aggressive & 2 & 2 & 2 & 2 & 1/300 & 2000 & \begin{minipage}{.3\textwidth}\includegraphics[width=\linewidth]{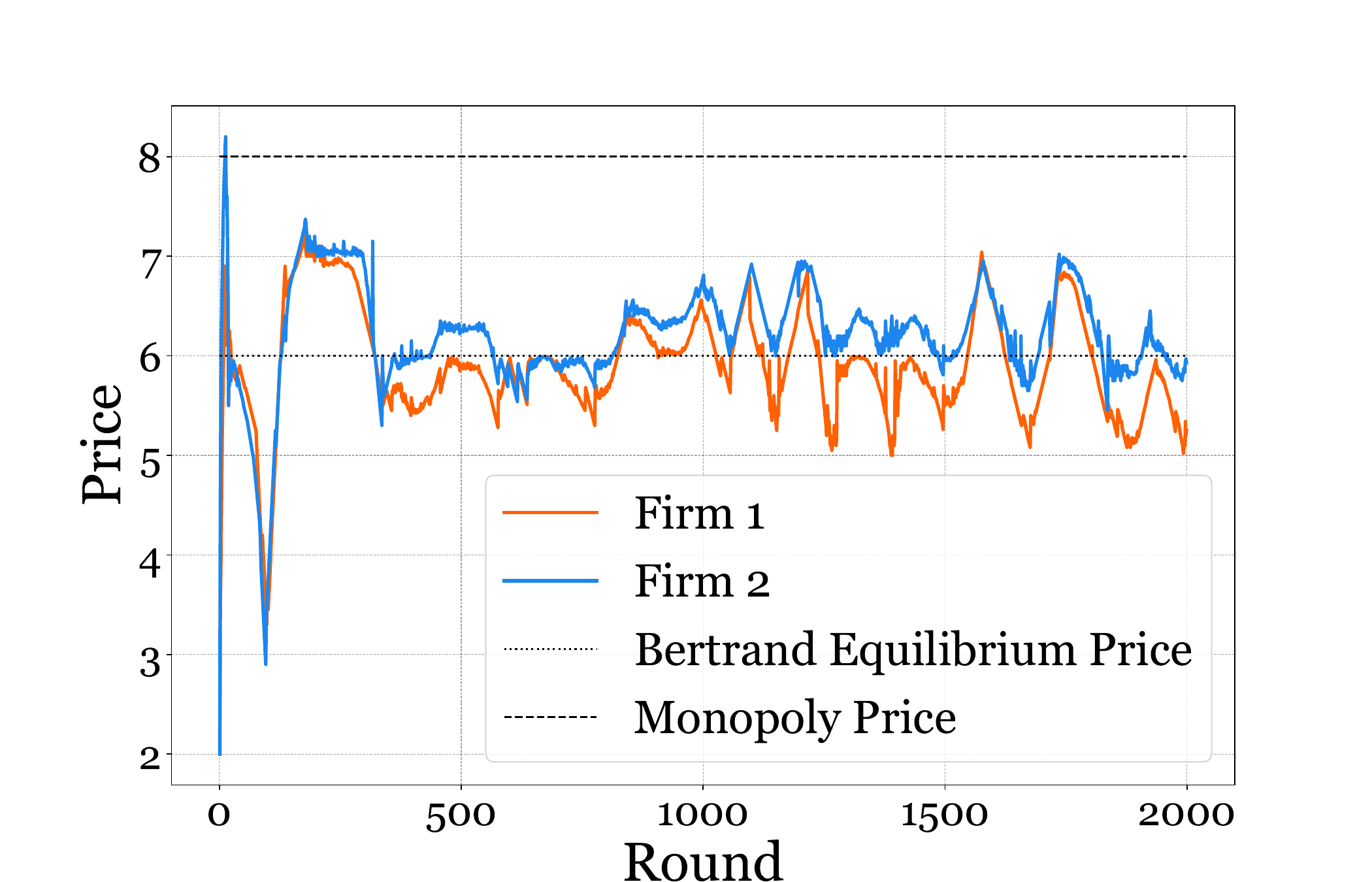}\end{minipage} \\ \hline
            4 & True & False & None & 2 & 2 & 2 & 2 & 1/300 & 400 & \begin{minipage}{.3\textwidth}\includegraphics[width=\linewidth]{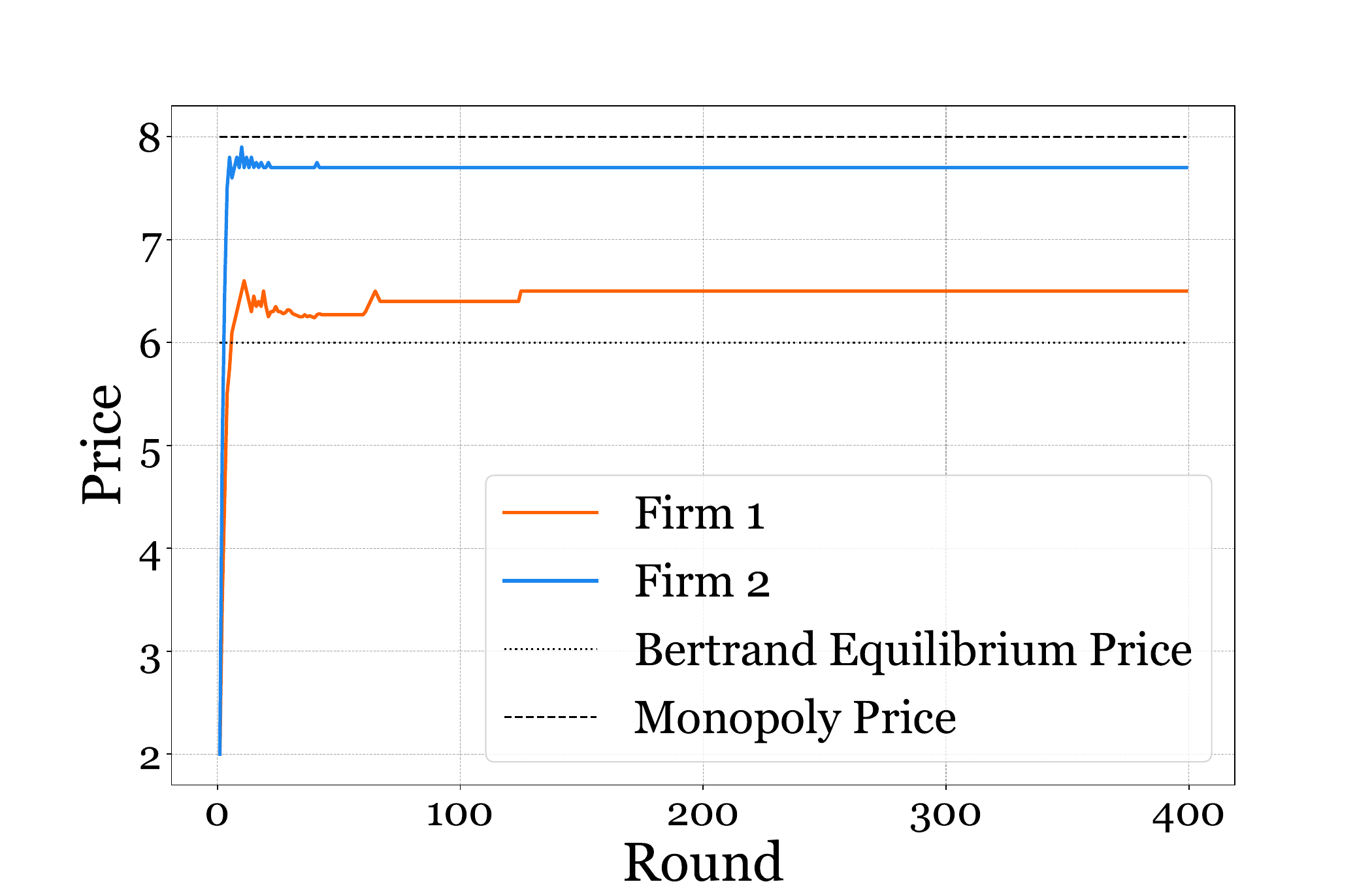}\end{minipage} \\
            \hline
        \end{tabular}
    }
    \label{tab:firm-competition-results-1}
\end{table*}

\begin{table*}[h]
    \centering
    \caption{Firm competition results. Group 2: Cost 1 $\neq$ Cost 2 (with planning without conversation).}
    \resizebox{\linewidth}{!}{
        \begin{tabular}{|c|c|c|c|c|c|c|c|c|c|c|}
            \hline
            \# & Plng & Cnvs & Psn & $c_1$ & $c_2$ & $i_1$ & $i_2$ & $d$ & $r$ & Figure \\
            \hline
            1 & True & False & Active & 2 & 5 & 2 & 5 & 1/300 & 1000 & \begin{minipage}{.3\textwidth}\includegraphics[width=\linewidth]{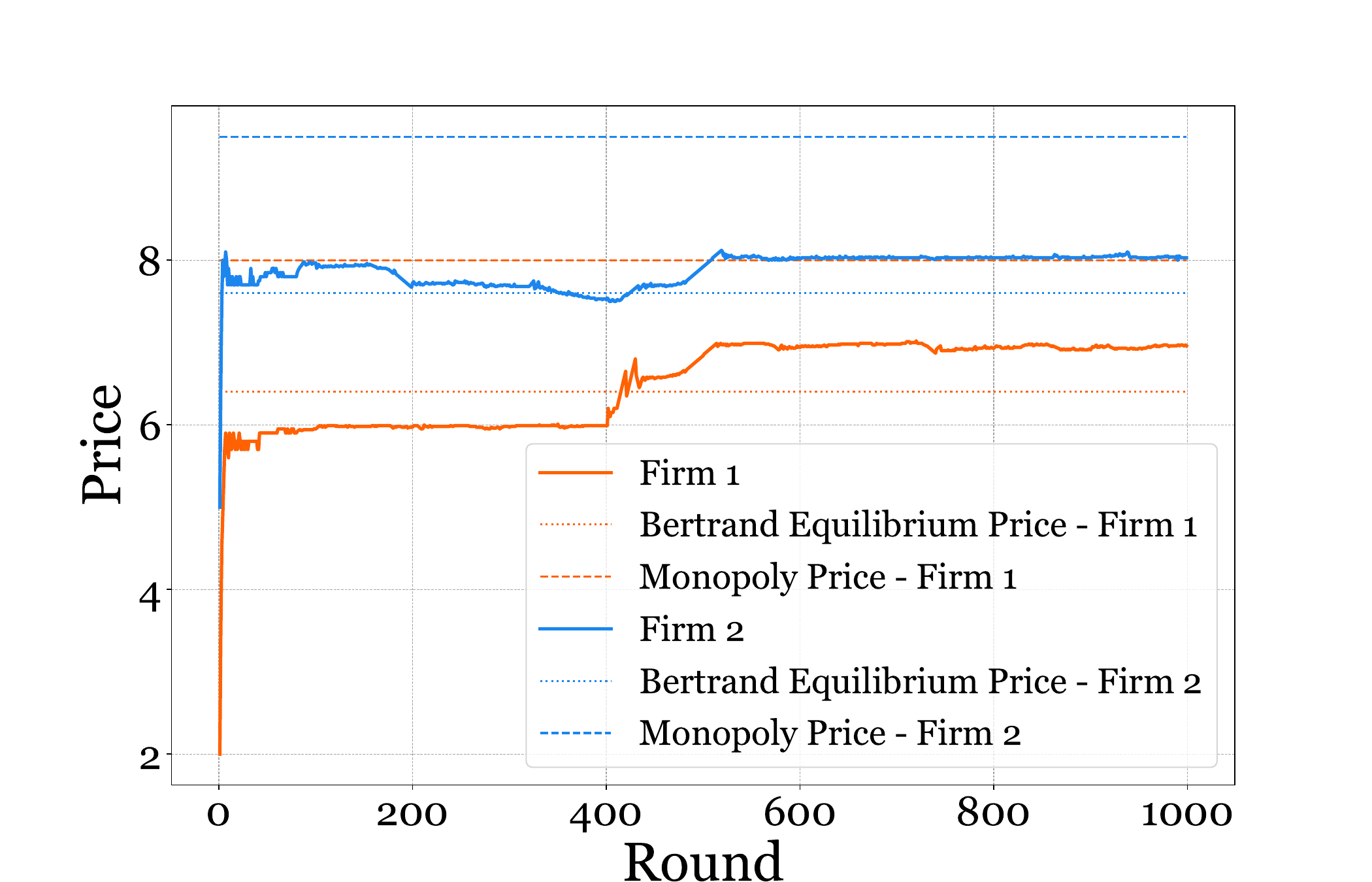}\end{minipage} \\ \hline
            2 & True & False & Aggressive & 2 & 5 & 10 & 10 & 1/300 & 600 & \begin{minipage}{.3\textwidth}\includegraphics[width=\linewidth]{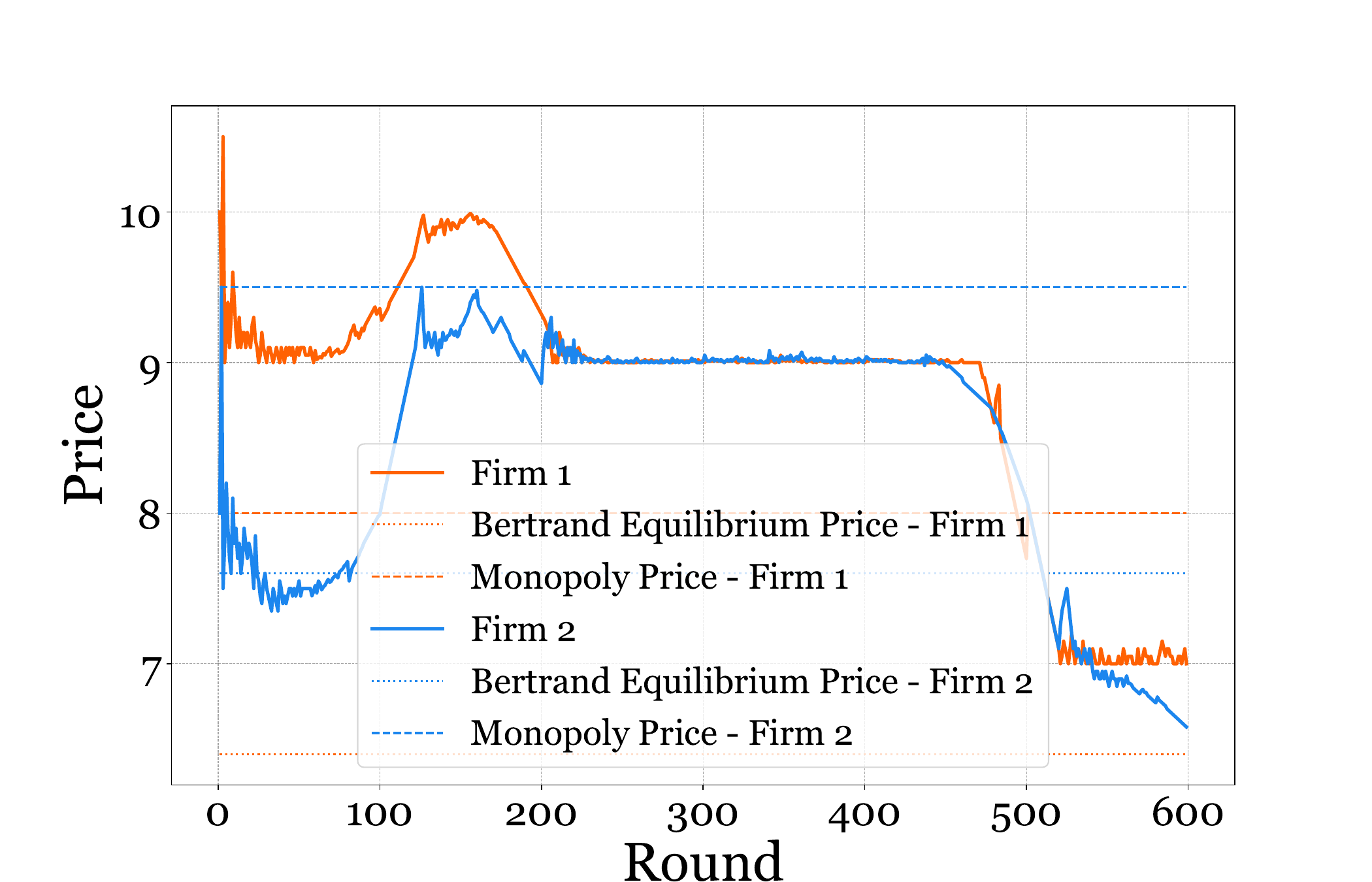}\end{minipage} \\ \hline
            3 & True & False & Aggressive & 2 & 5 & 2 & 5 & 1/300 & 1200 & \begin{minipage}{.3\textwidth}\includegraphics[width=\linewidth]{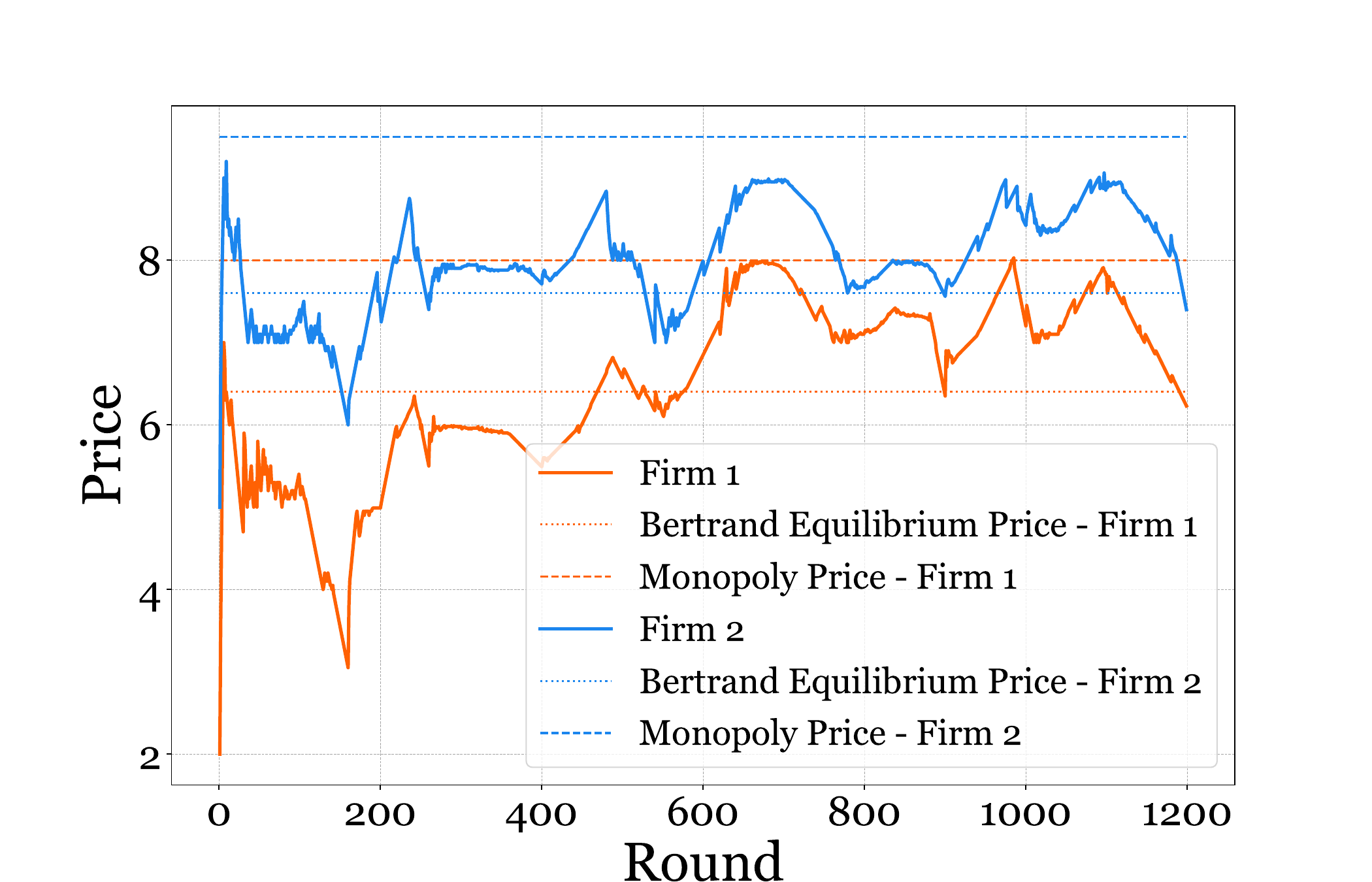}\end{minipage} \\
            \hline
        \end{tabular}
    }
    \label{tab:firm-competition-results-2}
\end{table*}

\begin{table*}[h]
    \centering
    \caption{Firm competition results. Group 3: $d = 0$ (with planning without conversation).}
    \resizebox{\linewidth}{!}{
        \begin{tabular}{|c|c|c|c|c|c|c|c|c|c|c|}
            \hline
            \# & Plng & Cnvs & Psn & $c_1$ & $c_2$ & $i_1$ & $i_2$ & $d$ & $r$ & Figure \\
            \hline
            1 & True & False & Active & 2 & 2 & 2 & 2 & 0 & 700 & \begin{minipage}{.3\textwidth}\includegraphics[width=\linewidth]{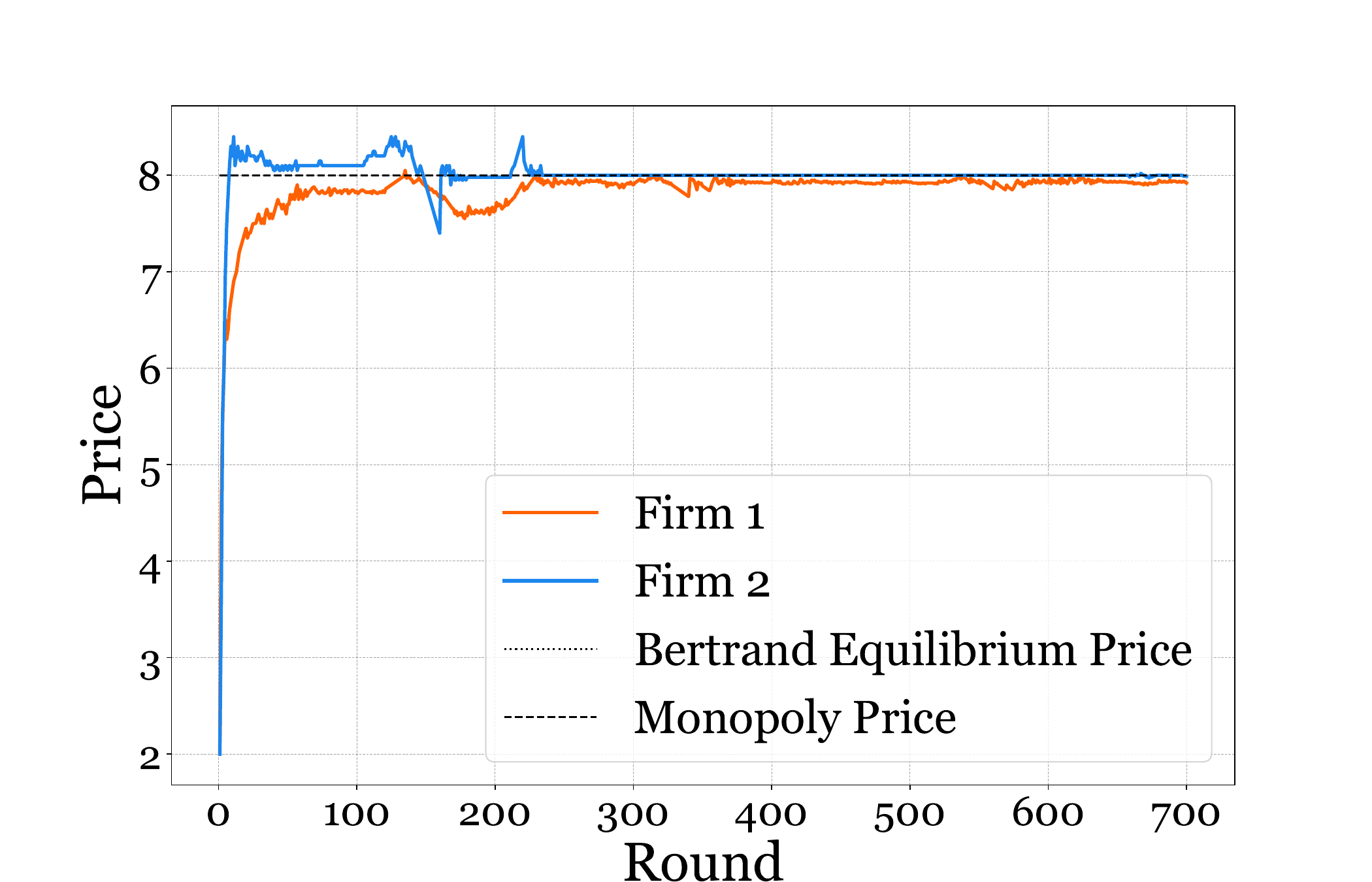}\end{minipage} \\ \hline
            2 & True & False & Aggressive & 2 & 2 & 2 & 2 & 0 & 800 & \begin{minipage}{.3\textwidth}\includegraphics[width=\linewidth]{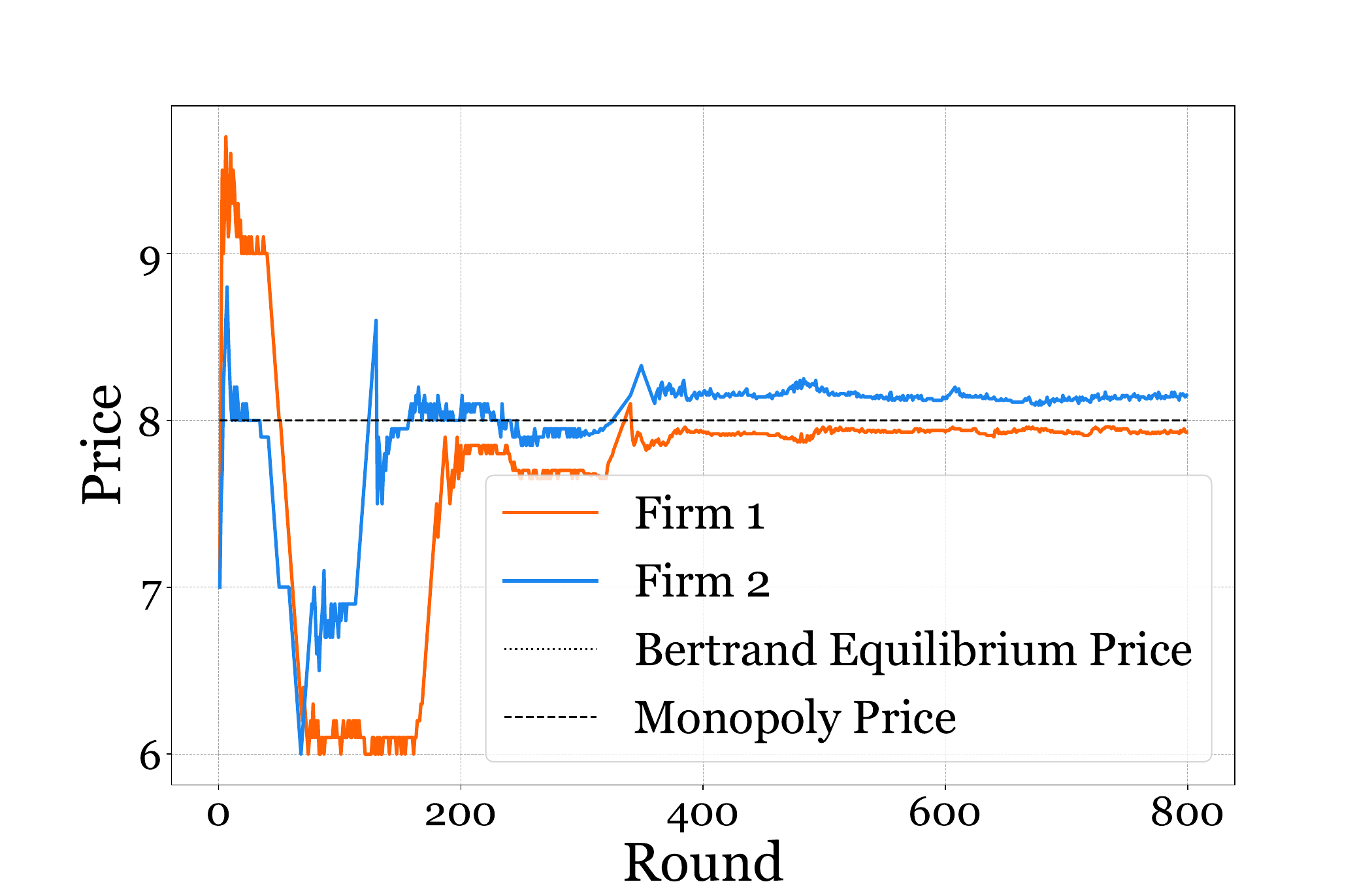}\end{minipage} \\
            \hline
        \end{tabular}
    }
    \label{tab:firm-competition-results-3}
\end{table*}

\begin{table*}[h]
    \centering
    \caption{Firm competition results. Group 4: $d = \beta = 1/300$ (with planning without conversation).}
    \resizebox{\linewidth}{!}{
        \begin{tabular}{|c|c|c|c|c|c|c|c|c|c|c|}
            \hline
            \# & Plng & Cnvs & Psn & $c_1$ & $c_2$ & $i_1$ & $i_2$ & $d$ & $r$ & Figure \\
            \hline
            1 & True & False & Active & 2 & 2 & 2 & 2 & 1/300 & 1800 & \begin{minipage}{.3\textwidth}\includegraphics[width=\linewidth]{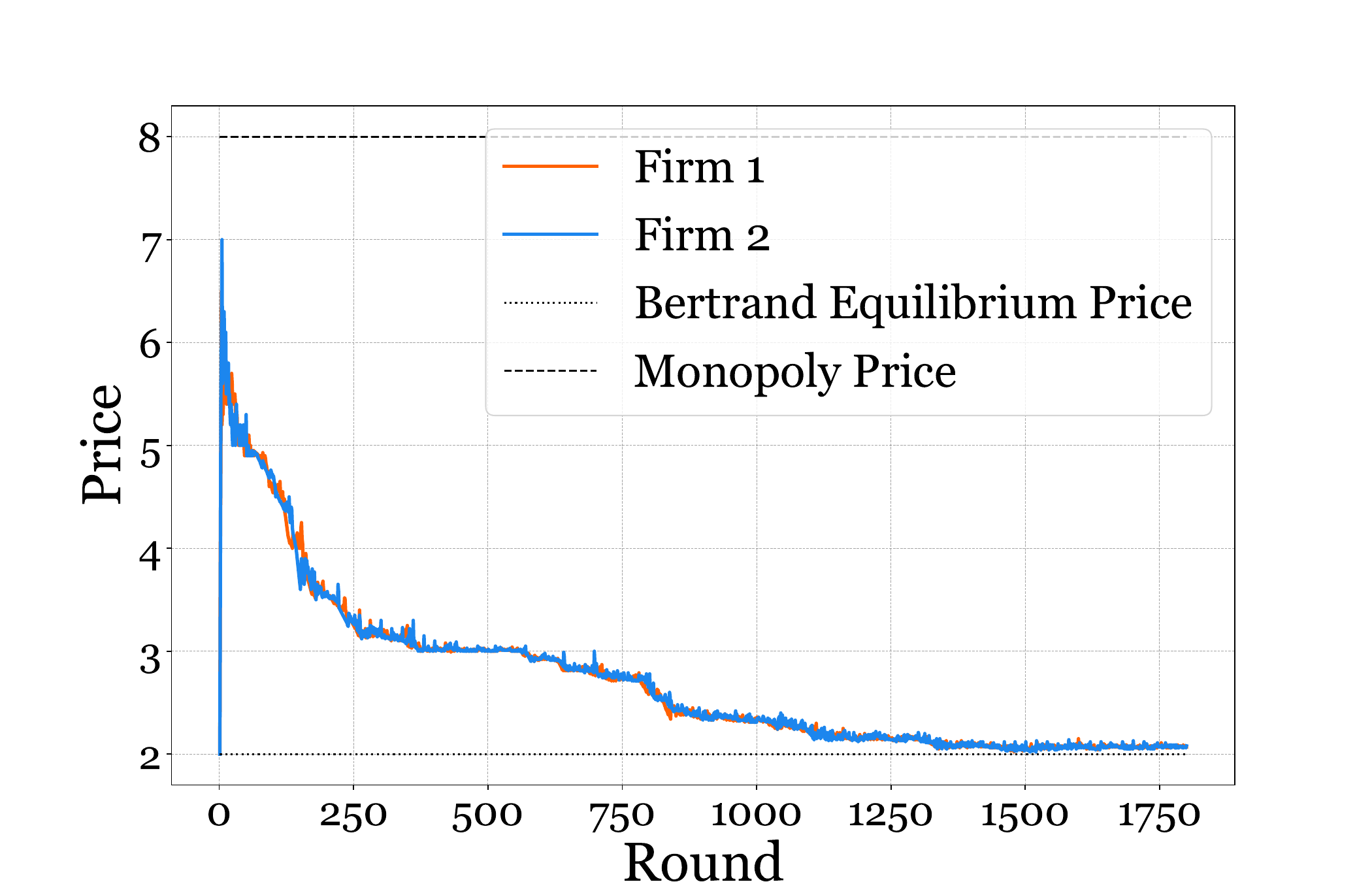}\end{minipage} \\
            \hline
        \end{tabular}
    }
    \label{tab:firm-competition-results-4}
\end{table*}

\begin{table*}[h]
    \centering
    \caption{Firm competition results. Group 5: Varying initial prices (with planning without conversation).}
    \resizebox{\linewidth}{!}{
        \begin{tabular}{|c|c|c|c|c|c|c|c|c|c|c|}
            \hline
            \# & Plng & Cnvs & Psn & $c_1$ & $c_2$ & $i_1$ & $i_2$ & $d$ & $r$ & Figure \\
            \hline
            1 & True & False & Active & 2 & 2 & 2 & 10 & 1/300 & 500 & \begin{minipage}{.3\textwidth}\includegraphics[width=\linewidth]{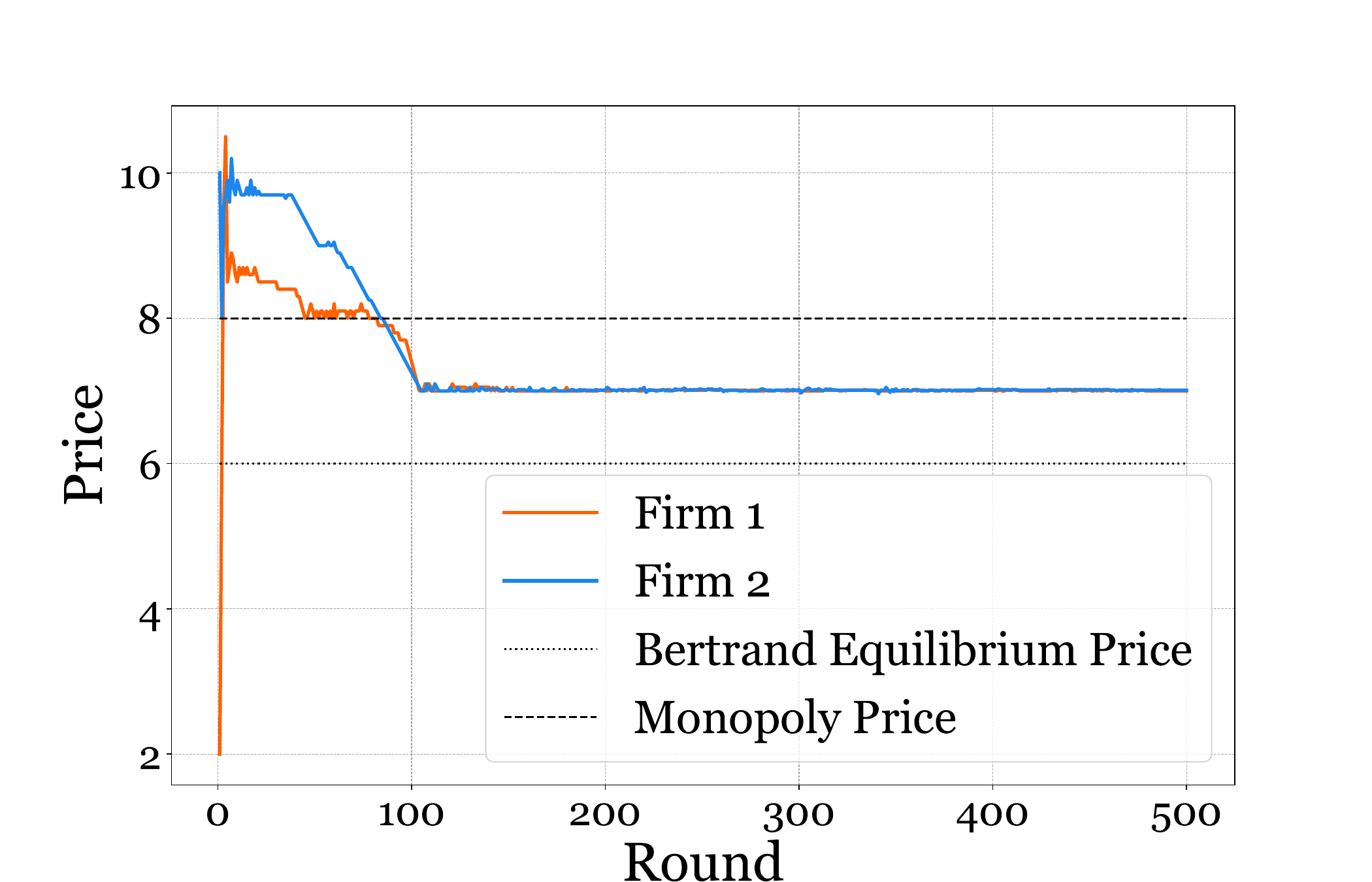}\end{minipage} \\ \hline
            2 & True & False & Active & 2 & 2 & 7 & 7 & 1/300 & 1000 & \begin{minipage}{.3\textwidth}\includegraphics[width=\linewidth]{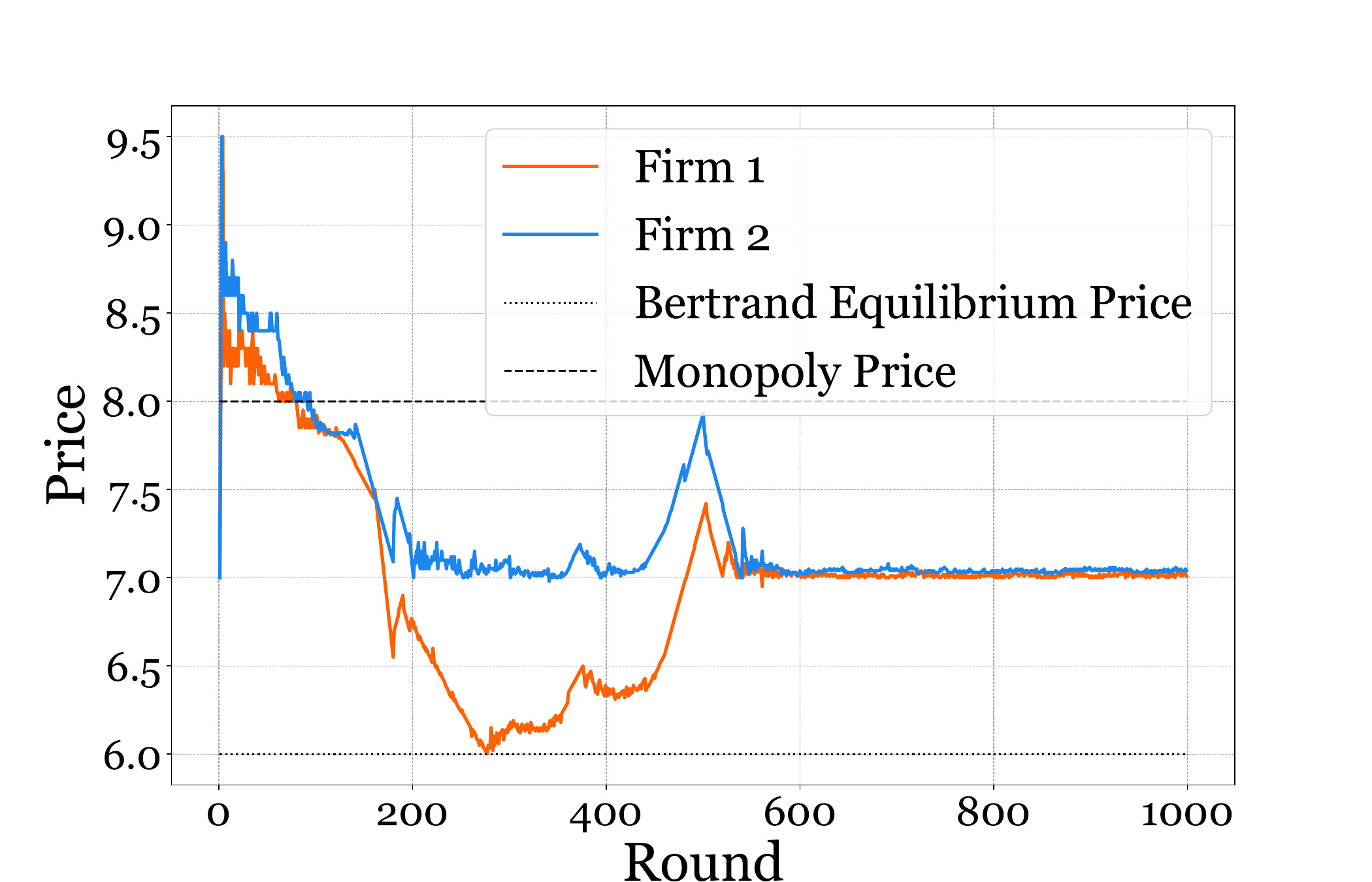}\end{minipage} \\
            \hline
        \end{tabular}
    }
    \label{tab:firm-competition-results-5}
\end{table*}

\begin{table*}[h]
    \centering
    \caption{Firm competition results. Group 6: Ablation study 1 -- effect of conversation (with planning with conversation).}
    \resizebox{\linewidth}{!}{
        \begin{tabular}{|c|c|c|c|c|c|c|c|c|c|c|}
            \hline
            \# & Plng & Cnvs & Psn & $c_1$ & $c_2$ & $i_1$ & $i_2$ & $d$ & $r$ & Figure \\
            \hline
            1 & True & True & Active & 2 & 2 & 2 & 2 & 1/300 & 1200 & \begin{minipage}{.3\textwidth}\includegraphics[width=\linewidth]{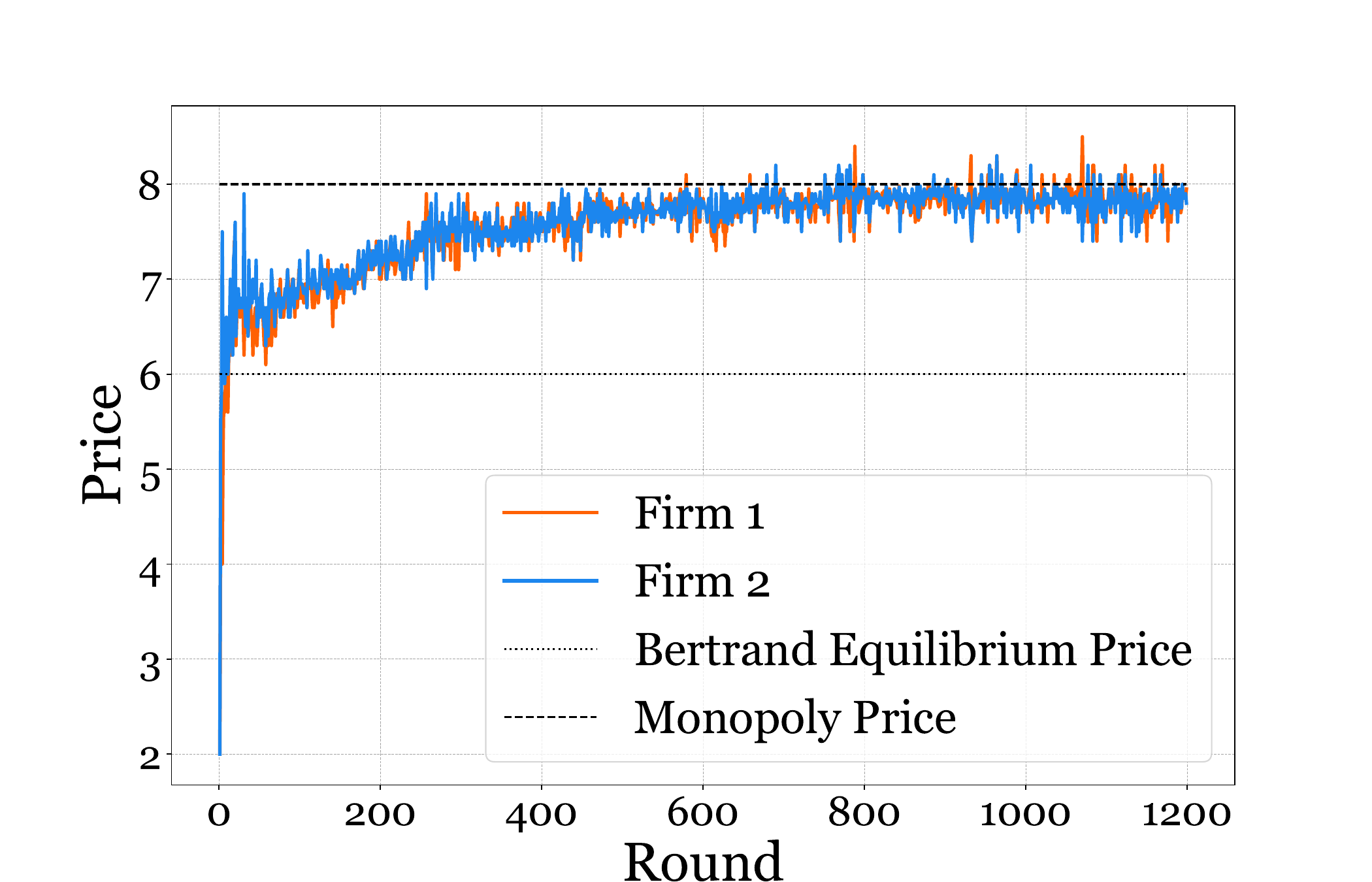}\end{minipage} \\
            \hline
        \end{tabular}
    }
    \label{tab:firm-competition-results-6}
\end{table*}

\begin{table*}[h]
    \centering
    \caption{Firm competition results. Group 7: Ablation study 2 -- shift from with conversation to without conversation, with prices of the first 400 rounds of Group 6-1.}
    \resizebox{\linewidth}{!}{
        \begin{tabular}{|c|c|c|c|c|c|c|c|c|c|c|}
            \hline
            \# & Plng & Cnvs & Psn & $c_1$ & $c_2$ & $i_1$ & $i_2$ & $d$ & $r$ & Figure \\
            \hline
            1 & True & True (400 rounds) + False (200 rounds) & Active & 2 & 2 & 2 & 2 & 1/300 & 600 & \begin{minipage}{.48\textwidth}\includegraphics[width=\linewidth]{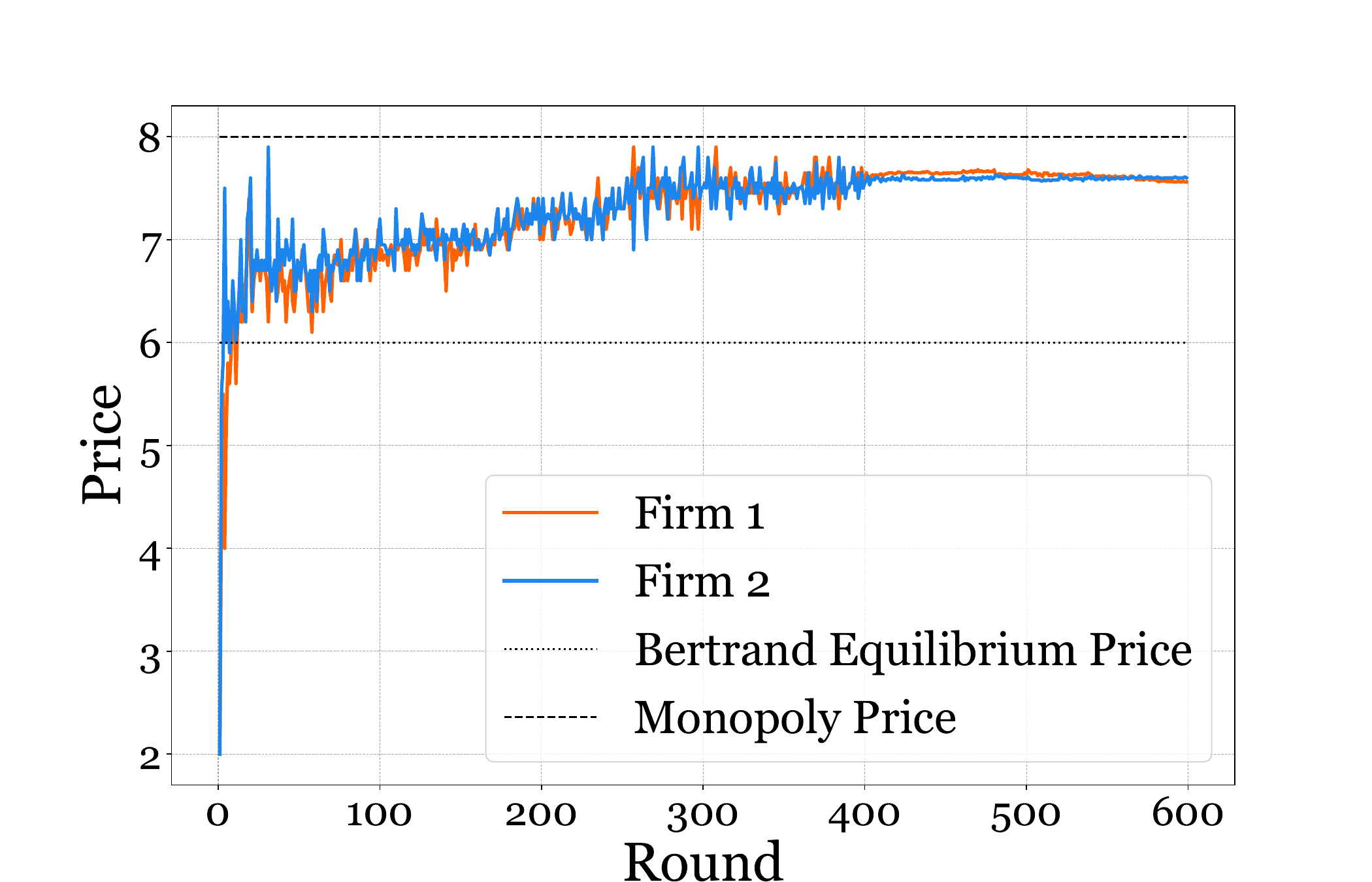}\end{minipage} \\
            \hline
        \end{tabular}
    }
    \label{tab:firm-competition-results-7}
\end{table*}

\begin{table*}[h]
    \centering
    \caption{Firm competition results. Group 8: Ablation study 3 -- effect of planning (without planning without conversation).}
    \resizebox{\linewidth}{!}{
        \begin{tabular}{|c|c|c|c|c|c|c|c|c|c|c|}
            \hline
            \# & Plng & Cnvs & Psn & $c_1$ & $c_2$ & $i_1$ & $i_2$ & $d$ & $r$ & Figure \\
            \hline
            1 & False & False & Active & 2 & 2 & 2 & 2 & 1/300 & 550 & \begin{minipage}{.3\textwidth}\includegraphics[width=\linewidth]{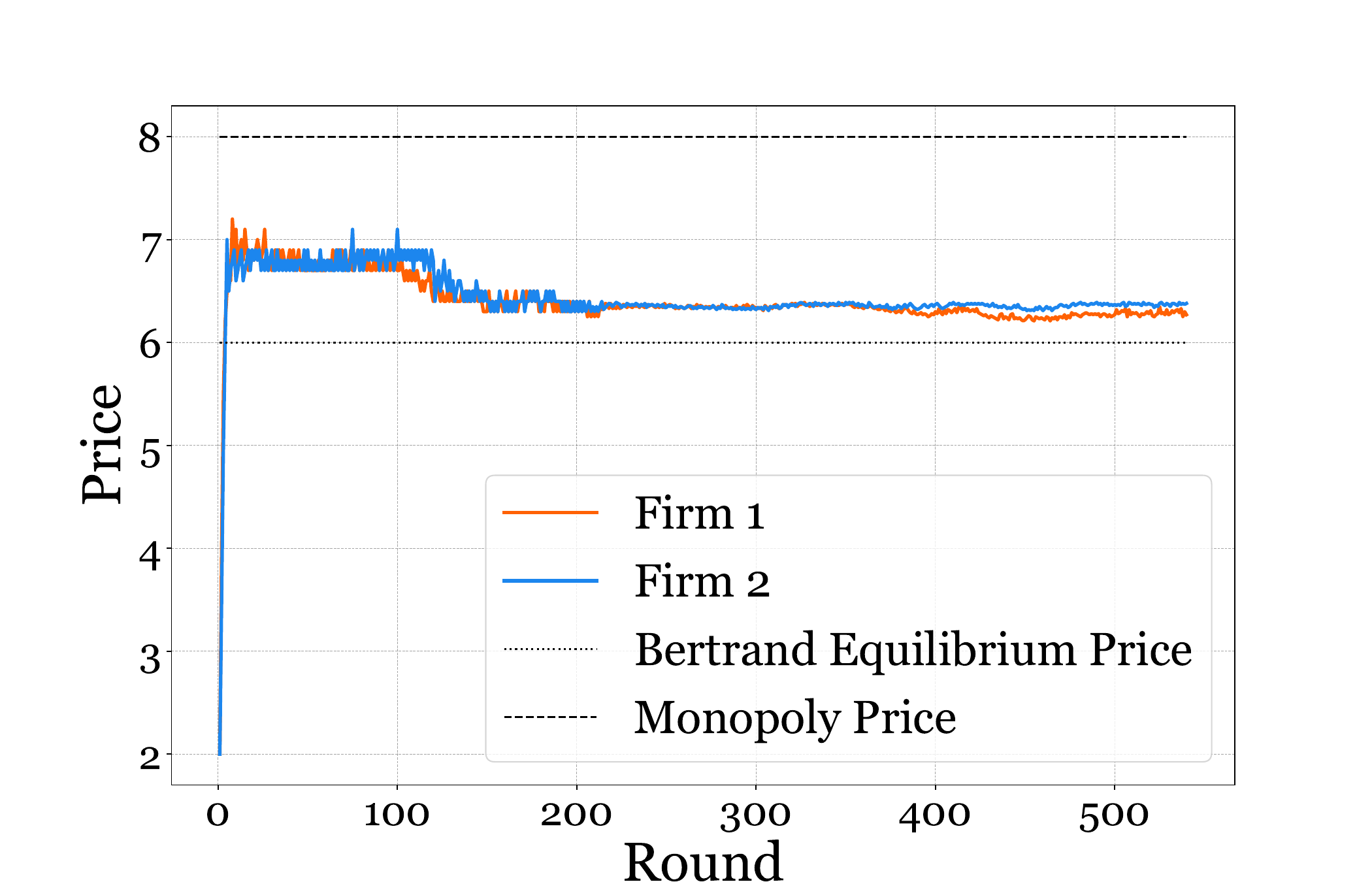}\end{minipage} \\ \hline
            2 & False & False & Aggressive & 2 & 2 & 2 & 2 & 1/300 & 1800 & \begin{minipage}{.3\textwidth}\includegraphics[width=\linewidth]{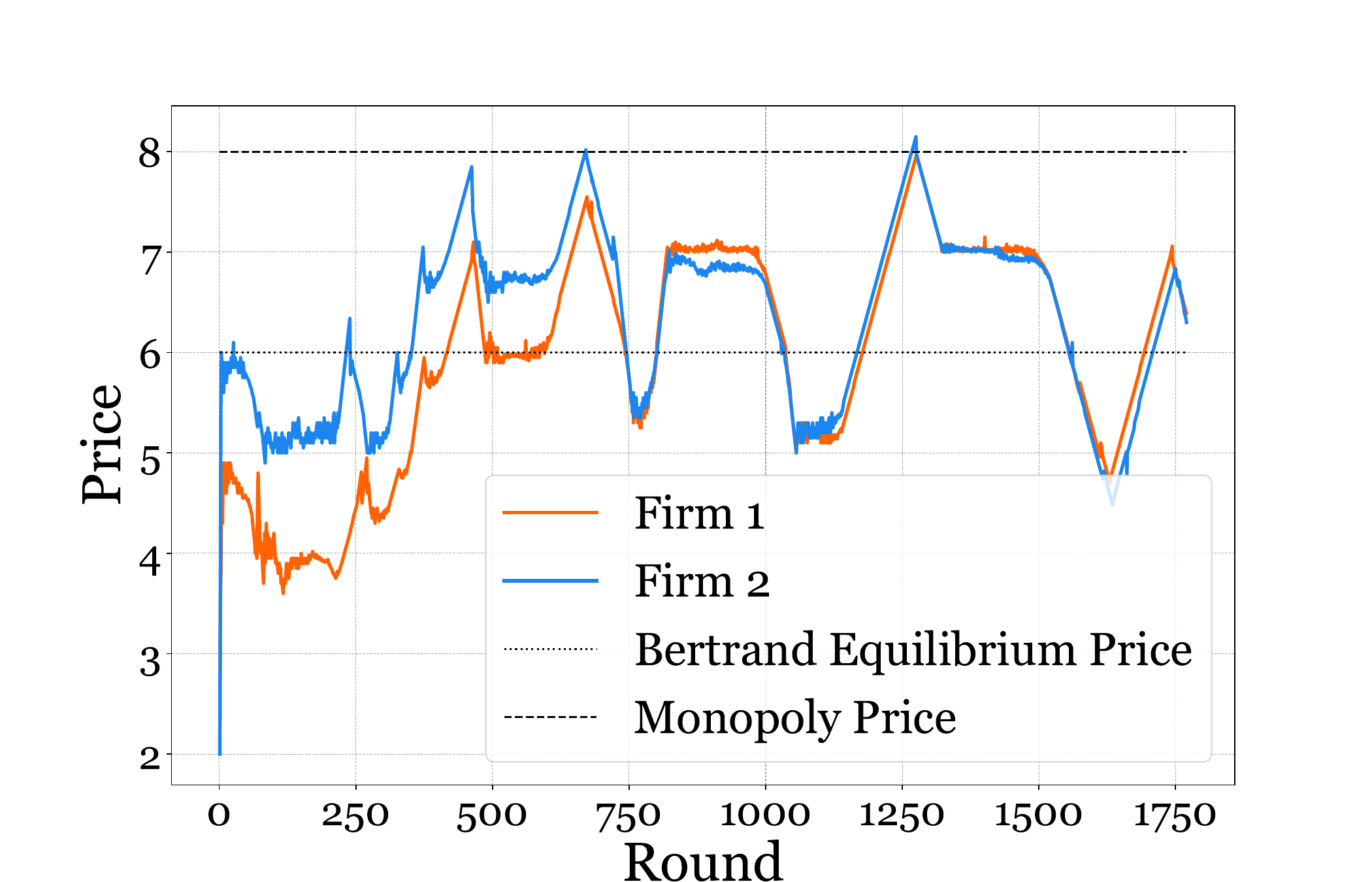}\end{minipage} \\
            \hline
        \end{tabular}
    }
    \label{tab:firm-competition-results-8}
\end{table*}

\begin{table*}[h]
    \centering
    \caption{Firm competition results. Group 9: Ablation study 4 -- shift from without planning to with planning, with prices of the first 100 rounds of Group 8-2.}
    \resizebox{\linewidth}{!}{
        \begin{tabular}{|c|c|c|c|c|c|c|c|c|c|c|}
            \hline
            \# & Plng & Cnvs & Psn & $c_1$ & $c_2$ & $i_1$ & $i_2$ & $d$ & $r$ & Figure \\
            \hline
            1 & False (100 rounds) + True (500 rounds) & False & Aggressive & 2 & 2 & 2 & 2 & 1/300 & 600 & \begin{minipage}{.48\textwidth}\includegraphics[width=\linewidth]{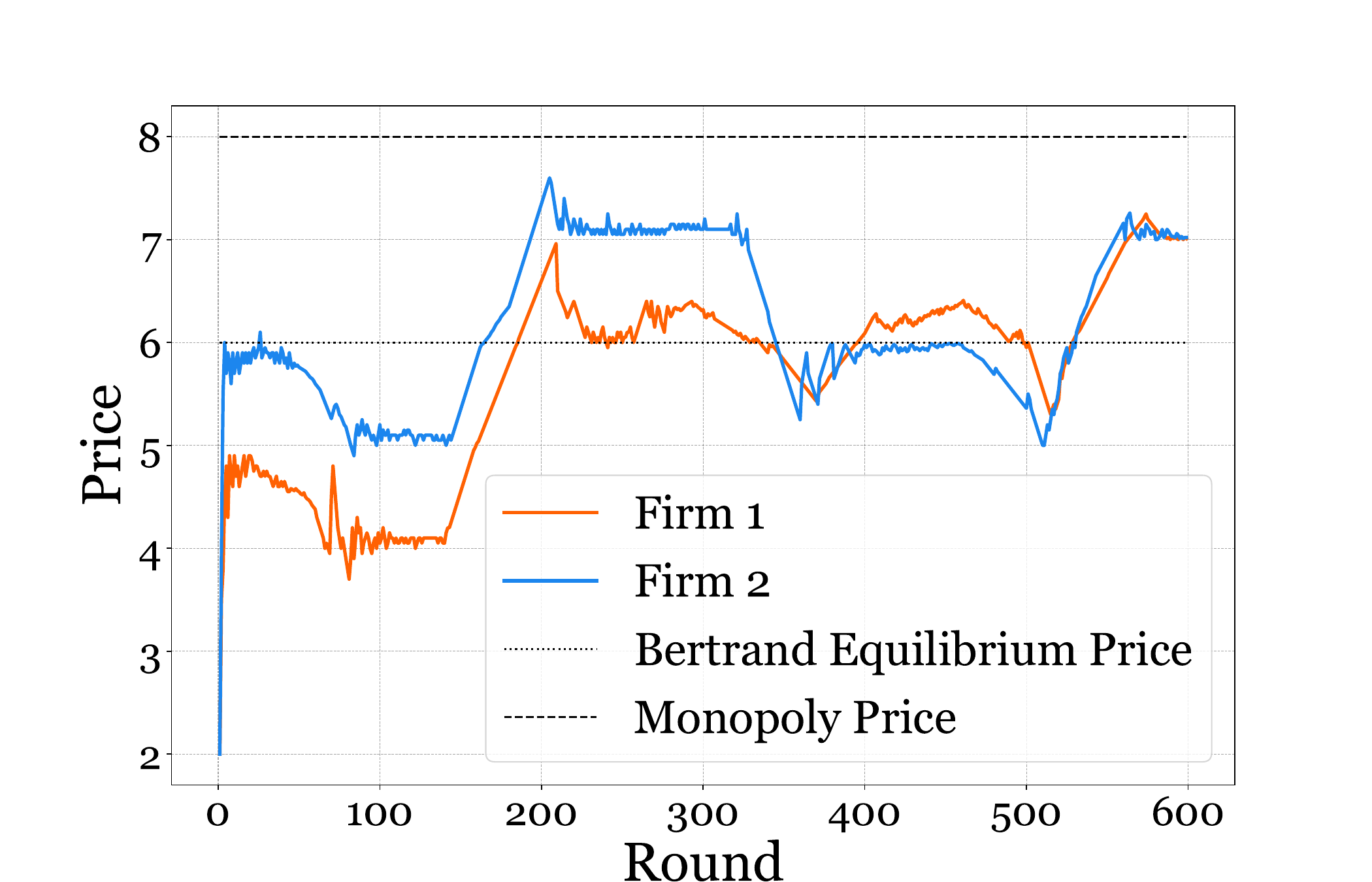}\end{minipage} \\
            \hline
        \end{tabular}
    }
    \label{tab:firm-competition-results-9}
\end{table*}

\end{document}